\documentclass{article}
\usepackage{arxiv}

\usepackage{geometry}
\usepackage{graphicx}
\usepackage[hidelinks]{hyperref}

\usepackage{amssymb}
\usepackage{bm}
\usepackage{lineno}
\usepackage{amsmath}

\usepackage{multirow}
\usepackage{float}
\usepackage{tabularx}
\usepackage{subcaption}

\usepackage{array,xcolor}

\usepackage{algorithm, algpseudocode}

\usepackage{cite}
\title{Randomized prior wavelet neural operator for uncertainty quantification}
\author{
  Shailesh Garg  \\
  Department of Applied Mechanics\\
  Indian Institute of Technology Delhi\\
  Hauz Khas, New Delhi 110016, India. \\
  \texttt{shaileshgarg96@gmail.com} \\
  \And
  Souvik Chakraborty  \\
  Department of Applied Mechanics\\
  Yardi School of Artificial Intelligence (YScAI)\\
  Indian Institute of Technology Delhi\\
  Hauz Khas, New Delhi 110016, India. \\
  \texttt{souvik@am.iitd.ac.in}}
\begin{document}
\maketitle
\begin{abstract}
In this paper, we propose a novel data-driven operator learning framework referred to as the \textit{Randomized Prior Wavelet Neural Operator} (RP-WNO). The proposed RP-WNO is an extension of the recently proposed wavelet neural operator, which boasts excellent generalizing capabilities but cannot estimate the uncertainty associated with its predictions. RP-WNO, unlike the vanilla WNO, comes with inherent uncertainty quantification module and hence, is expected to be extremely useful for scientists and engineers alike. RP-WNO utilize randomized prior networks, which can account for prior information and is easier to implement for large, complex deep-learning architectures than its Bayesian counterpart.
Four examples have been solved to test the proposed framework, and the results produced advocate favorably for the efficacy of the proposed framework.
\end{abstract}

\section{Introduction}\label{section: Introduction}
Uncertainty is an inextricable part of any dataset originating from a physical process. The physical process under consideration can be natural, like weather phenomena \cite{slingo2011uncertainty,palmer2005representing}, or artificial, like building deflections \cite{moon2006impacts,mace2005uncertainty,vakhshouri2014limitations}. These uncertainties \cite{der2009aleatory,rowe1994understanding} can manifest themselves in the form of (i) aleatoric uncertainty or (ii) epistemic uncertainty, or both. Aleatoric uncertainty is primarily present due to the inherent randomness in the physical process. Measurement noise can also be a potential source of aleatoric uncertainty in datasets collected using the internet of things. Since the sources of aleatoric uncertainty are ineradicable, it is often termed irreducible uncertainty. The other form of uncertainty that may be present is epistemic uncertainty, which originates from incomplete knowledge about the physical process under consideration. Since, theoretically, one can increase their understanding of a physical process by collecting sufficient data, epistemic uncertainty is also termed reducible uncertainty. Even though theoretically, the epistemic uncertainty is reducible given sufficient data, practically, it can never be entirely eliminated due to constraints like limited storage capacity and limited data collection time.

Now, while working with any physical process, knowing the associated uncertainty in various forms can lend the user auxiliary confidence, especially in decision-making tasks. Knowing aleatoric uncertainty can inform the user about the deviation of the system from a mean state, and knowing the epistemic uncertainty tells the user about the adequacy of the dataset chosen for decision-making. Let us now momentarily focus on the problem of generating datasets for any physical process. Due to physical constraints (like time and field storage capacity) and scientific constraints (like sensor technology or placement issues), the real-life data available for various physical processes is limited. Thus it falls upon the user to augment this data with artificial data to increase the total quantum. To generate this artificial data, we need to develop suitable computationally efficient algorithms that can learn the mapping between a physical system's observed inputs and outputs. Furthermore, in applications like reliability analysis \cite{rackwitz2001reliability,garg2022assessment,chakraborty2015semi} or design optimization \cite{rao2012mechanical,chatterjee2019critical,chakraborty2017surrogate}, where repeated simulations are needed, these algorithms can be used as a surrogate model to reduce the computational cost associated with detailed analysis.

In the available literature, techniques like polynomial chaos expansion \cite{kersaudy2015new,novak2018polynomial,sudret2014polynomial}, response surface method \cite{zhang2017time,lu2018improved,gogu2013efficient,chakraborty2014adaptive,alibrandi2014response}, and Gaussian process \cite{su2017gaussian,zhu2021efficient,chakraborty2021role,dubourg2011reliability} have been used to develop surrogate models in the past, but in recent times, researchers have started to explore heavily deep learning-based techniques. Deep learning \cite{lecun2015deep,deng2014deep} algorithms have gained traction in a wide array of fields \cite{shinde2018review,najafabadi2015deep,ker2017deep,mosavi2019state,ardabili2020covid,daniell2022physics,rahman2022leveraging,maulik2020probabilistic,maulik2017neural} owing to their enormous capacity for modification, low prediction cost, and flexibility in implementation. Learning the mapping between inputs and outputs of a physical process can now be achieved by deep learning based data-driven operator learning algorithms. Recently proposed Wavelet Neural Operator \cite{tripura2022wavelet,thakur2022multi} (WNO), DeepONet \cite{lu2021learning,yang2022scalable,garg2022assessment}, and Fourier Neural Operator \cite{li2020fourier} are among the popular operator learning algorithms which have shown tremendous generalization capacity despite being data-driven algorithms. Function approximation algorithms like Physics Informed Neural Networks \cite{raissi2019physics}, which utilize the system's governing equation directly, may also be used. However, these are limited in their applications as complete knowledge of physics is required to converge successfully to the solution. Such limitations are not placed on data-driven algorithms.

Theoretically, estimating aleatoric uncertainty is trivial given a sufficiently large dataset, which is feasible using an appropriate surrogate model. However, estimating epistemic uncertainty is nontrivial and requires algorithms specifically designed for the same. In deep learning, uncertainty quantification can be achieved using either Bayesian or frequentist approaches. The former utilizes Bayesian inference \cite{box2011bayesian} for uncertainty quantification. Among Bayesian methods, Monte Carlo based approaches \cite{niu2012short,papamarkou2022challenges} are the most accurate, but they are impractical in the face of complex deep neural network architectures. Another popular Bayesian approach is to utilize variational inference \cite{graves2011practical,blundell2015weight} based methods. These, although can be implemented for complex neural architectures, give only an approximate solution and may fail to converge altogether for multi-modal posterior distributions.
Frequentist approaches, on the other hand, are generally easier to implement owing to them being set in a deterministic framework. In frequentist approaches, ensemble training \cite{rahaman2021uncertainty} is among the most popular way to estimate uncertainty. However, in vanilla ensemble training, there is no mechanism to include prior information.

In this paper, we introduce a novel Randomized Prior Wavelet Neural Operator (RP-WNO). The proposed RP-WNO utilizes the concepts of random prior networks \cite{osband2018randomized} and enables estimation of epistemic uncertainty (predictive uncertainty) associated with its predictions; this allows users to take well-informed decisions. The capacity of randomized prior networks to incorporate prior information while being set in a frequentist domain makes them a prime candidate for the proposed framework. Four examples have been covered in the paper showcasing the efficacy of the proposed algorithm in both maintaining the generalization capacity of WNO and enabling uncertainty quantification, which was previously not possible within the WNO architecture.

The rest of the paper is arranged as follows. Section \ref{section: proposed framework} discusses the WNO and introduces the proposed RP-WNO. Section \ref{section: Numerical Illustrations} discusses various examples showcasing the efficacy of the proposed framework, and section \ref{section: conclusion} concludes the manuscript. 
\section{Proposed framework}\label{section: proposed framework}
An  operator is a mathematical entity that transforms an $\mathcal I$ dimensional input function space to a $\mathcal O$ dimensional output function space. The relation can be described as,
\begin{equation}
    f_\mathcal I\in\mathbb R^\mathcal I\xrightarrow{\,\,\,\,\,\mbox{\normalsize$\mathbb O$}\,\,\,\,\,}f_\mathcal O\in\mathbb R^\mathcal O,
\end{equation}
where $\mathbb O$ denotes the operator, mapping the input functions to the output functions. In deep learning based operator learning schemes, the goal is to learn the operator by learning the mapping between the two function spaces. In WNO, wavelet transform is leveraged along with CNNs to learn the mapping and, by extension, the operator $\mathbb O$. 
\subsection{Wavelet neural operator}\label{subsection: wavelet neural operator}
WNO, proposed in \cite{tripura2022wavelet}, is a deep learning based operator learning scheme that utilizes wavelet blocks within its architecture to achieve its target. The effectiveness of any deep learning based scheme is heavily affected by the quality of data it receives for training. Data with concise and dense information is easier to learn when compared with data having a lot of noise and redundant information. This is simply explained by the fact that to learn a dense dataset, fewer weights and biases will be required, which makes the backpropagation easier and more precise. To this end, in WNO, wavelet transform is used, which extracts the relevant information from the input data while reducing the dimensionality, thus making it easier to learn the mapping between inputs and outputs.

WNO takes $n$ input functions along with the grid on which they are discretized as its inputs and lifts them to a higher dimension using a densely connected layer. For a two-dimensional domain, discretized using a grid having $r$ rows, $c$ columns, and $s$ samples, the above transaction in terms of dataset dimension evolution can be visualized as,
\begin{equation}
    \mathbf I\in\mathbb R^{s\times r\times c\times n+2}\xrightarrow{\,\,\,\,\,\mbox{\normalsize Dense layer}\,\,\,\,\,}\mathbf O_D\in\mathbb R^{s\times r\times c\times p},
\end{equation}
where $\mathbf I$ is the input to WNO and $\mathbf O_D$ is the output from the densely connected layer having $p$ nodes. Note that $p>n+2$ and in the absence of grid information for the input function, the input dimension will change from $\left(s\times r\times c\times n+2\right)$ to $\left(s\times r\times c\times n\right)$. The output from the densely connected layer is passed onto the wavelet block, and its dimensions are preserved after the final operations of the wavelet block. 
\subsubsection{Wavelet block}
In the WNO architecture, once the input is uplifted, it is passed onto wavelet blocks (WB) which are structured in such a way that they can extract relevant information from the data while keeping the parameter requirement of the neural network low. This is achieved by using wavelet transform and convolutional neural networks. WBs have two main components, (i) Wavelet Mapping Component (WMC) and (ii) Convolution Mapping Component (CMC). Both the components accept the same input and, after processing, return outputs with the same dimensions as input.

Wavelet Transform (WT) used in WMC divides the data into high-frequency and low-frequency wavelet coefficients and downsizes the data, extracting the relevant information. WMC first performs multi-level WT of the input. Through this, we can (i) distill only the relevant information while reducing the overall dimension and (ii) reduce the number of parameters required for learning features from the data. Upon satisfactory dimension reduction, the wavelet coefficients obtained after the final level of wavelet transform can be learned using simple dense linear layers. Now, using the output from the linear layers, inverse wavelet transform is performed. It should be noted that after the final level of the wavelet transform, either all or only the selected wavelet coefficients can be learned using the dense linear layers. The coefficients that are not learned are used as is while inverse wavelet transform.

The WT performed within the WMC can be continuous or discrete depending on the requirement of the dataset being learned. Continuous wavelet transform is more precise than the discrete wavelet transform, and it produces more wavelet coefficients that can be learned for better performance; however, it is computationally more expensive as compared to the discrete wavelet transform. The evolution of dataset size for the previous example of a two-dimensional dataset inside WMC, considering Discrete WT (DWT), can be tracked as follows,
\begin{multline}
    \mathbf I_{WB}\in\mathbb R^{s\times r\times c\times p}\xrightarrow{\,\,\,\,\,\mbox{\normalsize WT}\,\,\,\,\,}\mathbf O_{WT}\in\mathbb R^{s\times r'\times c'\times p\times 4}\xrightarrow{\,\,\,\,\,\mbox{\normalsize Dense linear layer}\,\,\,\,\,}\\\mathbf O_{DLL}\in\mathbb R^{s\times r'\times c'\times p\times 4}\xrightarrow{\,\,\,\,\,\mbox{\normalsize Inverse WT}\,\,\,\,\,}\mathbf O_{WMC}\in\mathbb R^{s\times r\times c\times p},
\end{multline}
where $I_{WB}$ is the input for the WB and $I_{WB} = \mathbf O_D$ for the first wavelet block inside WNO architecture. $\mathbf O_{WT}$ is output after performing multiple WTs. $r'<r$ and $s'<s$ are the reduced dimensions of the data after multi-level DWT, and since DWT will give four wavelet coefficients for the two-dimensional domain, the same is reflected in the dataset size. $\mathbf O_{DLL}$ is the output from the dense linear layer, which preserves its input size, and $\mathbf O_{WMC}$ is the final output obtained after the inverse wavelet transform. It should be noted that the dimension for the $\mathbf O_{WMC}$ is the same as that of the $\mathbf I_{WMC}$. 

\begin{figure}[ht!]
    \centering
    \includegraphics[width = \textwidth]{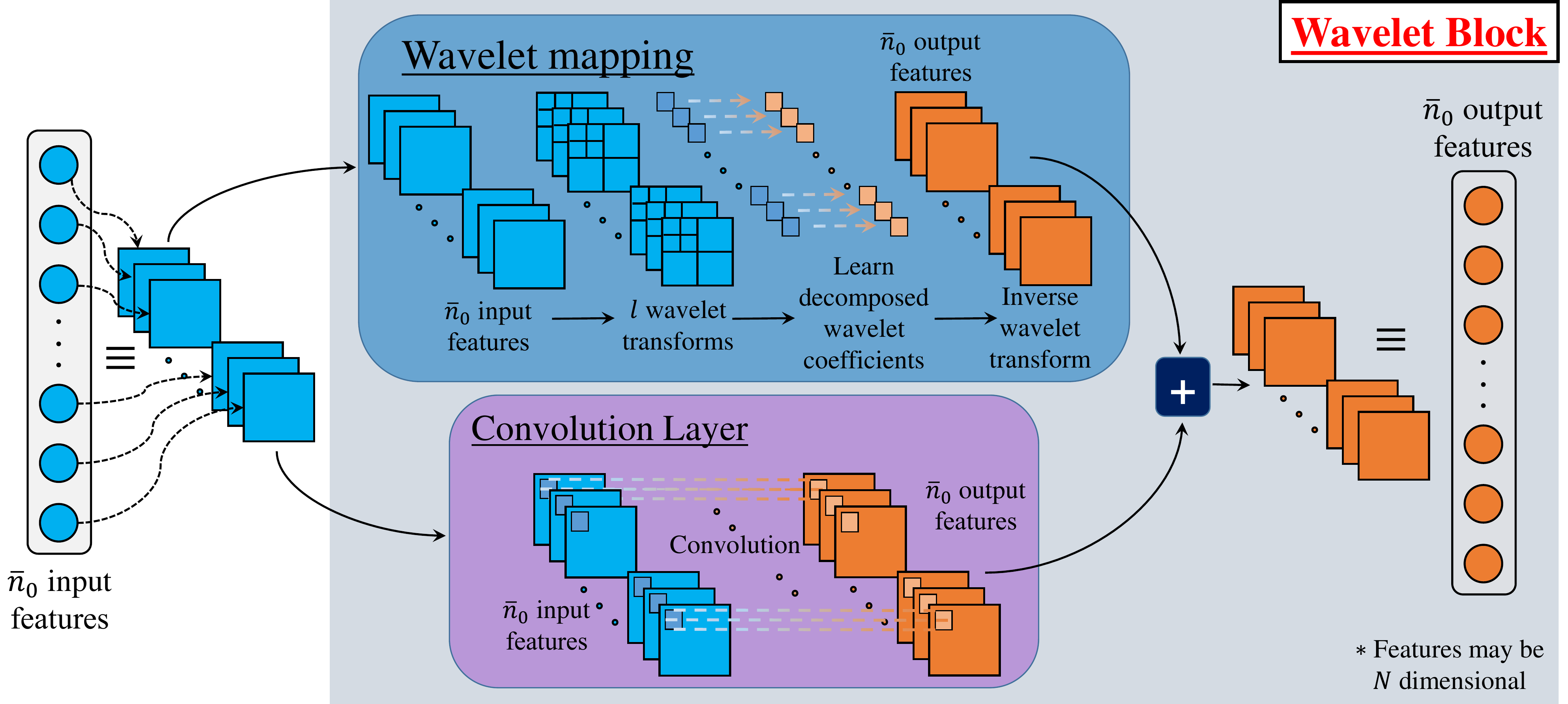}
    \caption{Schematic for information flow inside WBs.}
    \label{fig: wb}
\end{figure}

\begin{figure}[ht!]
    \centering
    \includegraphics[width = \textwidth]{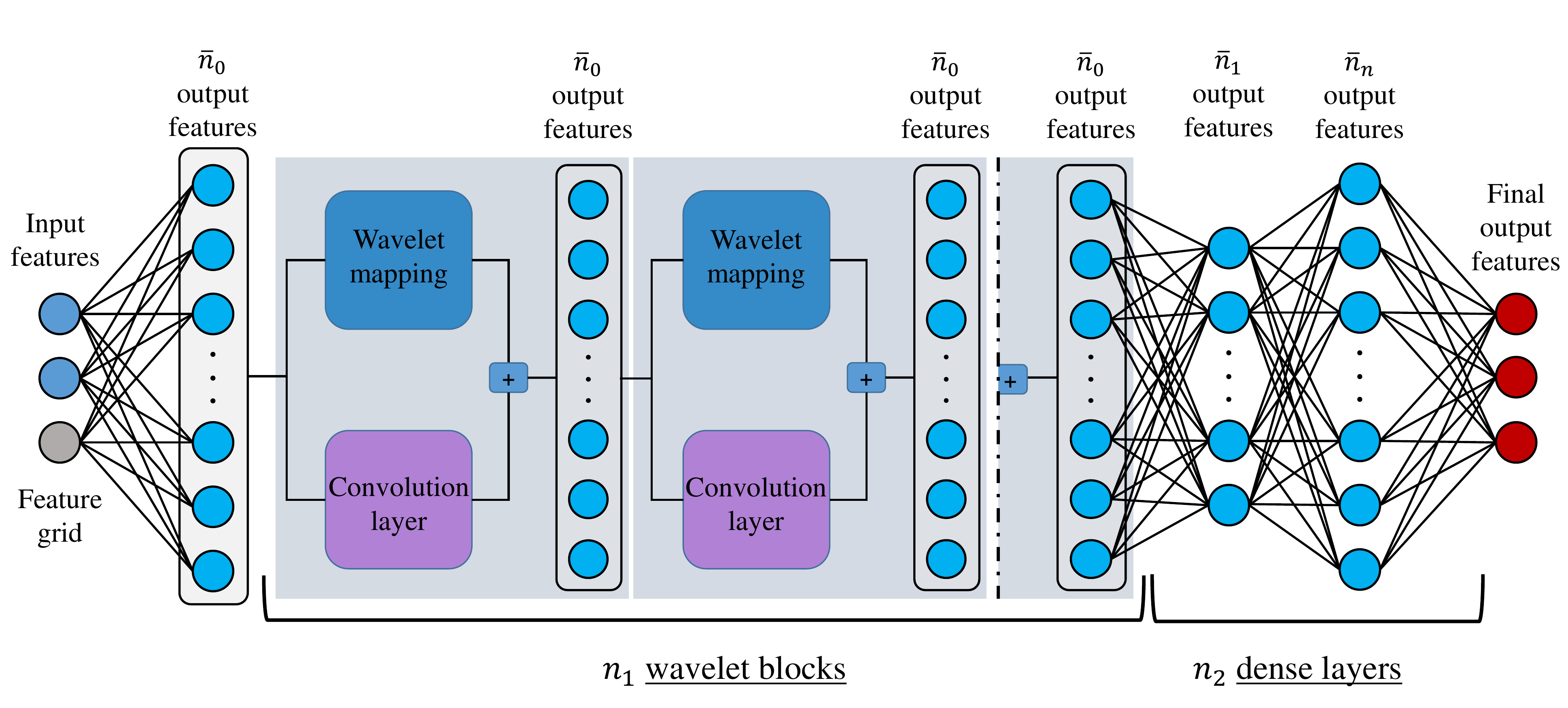}
    \caption{Schematic for the wavelet neural operator.}
    \label{fig: WNO}
\end{figure}

Next, inside CMC, convolution layers are used in such a way that the final output size is preserved. For the two-dimensional space, the evolution of dataset size inside CMC can be tracked as follows,
\begin{equation}
    \mathbf I_{WB}\in\mathbb R^{s\times r\times c\times p}\xrightarrow{\,\,\,\,\,\mbox{\normalsize Convolution layers}\,\,\,\,\,}\mathbf O_{CMC}\in\mathbb R^{s\times r\times c\times p},
\end{equation}
where $\mathbf O_{CMC}$ is the final output from the CMC. The outputs from the two components of WB are added in order to obtain the final output $\mathbf O_{WB} = \mathbf O_{WMC}+\mathbf O_{CMC}$. For second and subsequent WBs inside the WNO architecture, $\mathbf I_{WB}$ is equal to $\mathbf O_{WB}$ from the previous WB. A schematic for the information flow inside WB is shown in Fig. \ref{fig: wb}
After obtaining the output $\mathbf O_{WB}$, from the final wavelet block, densely connected layers are used in order to reduce the dimension of the uplifted data so that it may match the final output size. For the two-dimensional space, if the mapping is carried out for a single output function, the evolution of dataset size after the final wavelet block can be tracked as, 
\begin{equation}
    \mathbf O_{FWB}\in\mathbb R^{s\times r\times c\times p}\xrightarrow{\,\,\,\,\,\mbox{\normalsize Forward neural net}\,\,\,\,\,}\mathbf O_{WNO}\in\mathbb R^{s\times r\times c\times 1},
\end{equation}
where $\mathbf O_{FWB}$ is output of the final WB and $\mathbf O_{WNO}$ is the final output from the WNO. A schematic for the WNO architecture is shown in Fig. \ref{fig: WNO}.
For more details regarding WNO architecture, the readers are referred to \cite{tripura2022wavelet}.
\subsection{Randomized Prior WNO (RP-WNO)}\label{subsection: randomized prior wno}
The proposed RP-WNO utilizes Randomized Prior (RP) networks introduced in \cite{osband2018randomized} for uncertainty quantification.
RP networks utilize two neural networks with identical architectures but different random initialization to enable uncertainty quantification in deep learning models. Among the two networks, one has trainable weights and biases (trainable network), and the other has fixed weights and biases (non-trainable network). Similar to ensemble training, multiple copies of the randomized prior network are trained for uncertainty quantification. The advantage here is that the network with fixed weights and biases makes it possible to incorporate prior information, which was previously not possible in the vanilla ensemble training. Now, during training, in each forward pass, the output from the trainable and non-trainable networks are added before back-propagating for the trainable network. The output $\mathbf O_{RP}$ of a RP network can be described as,
\begin{equation}
    \mathbf O_{RP} = \mathbf O_T+\beta\mathbf O_{NT},
    \label{eq: RPN}
\end{equation}
where $O_T$ is the output of the trainable network and $\mathbf O_{NT}$ is the output for the non-trainable network. It should be noted that the loss function for the RP network is computed based on output obtained from Eq. \eqref{eq: RPN}, and while back-propagating, the weights and biases of only the trainable network are updated. $\beta$ in Eq. \eqref{eq: RPN} is a hyper-parameter and its value is user defined. In the prevailing literature, it is observed that $\beta = 1$ produces optimum results.

\begin{figure}[ht!]
    \centering
    \includegraphics[width = \textwidth]{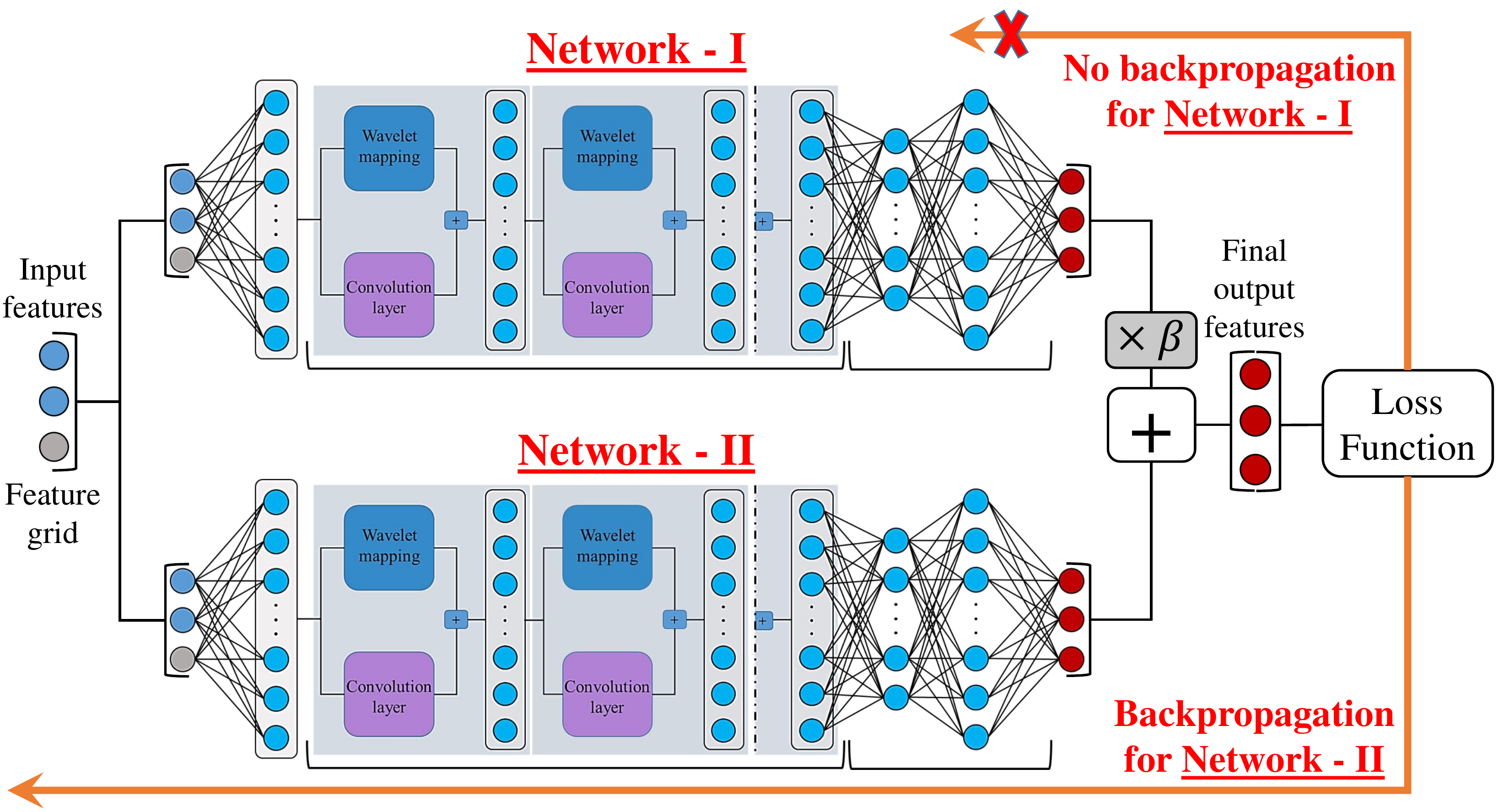}
    \caption{Schematic of the proposed RP-WNO.}
    \label{fig: RP-WNO}
\end{figure}

In the proposed RP-WNO, following the theory of RP networks, two identical WNO networks are taken, one being trainable and the other non-trainable. The outputs for the two are added, and the RP-WNO output $\mathbf O_{RP-WNO}$ is described as follows,
\begin{equation}
    \mathbf O_{RP-WNO} = \mathbf O_{WNOT}+\mathbf O_{WNONT},
    \label{eq: RP-WNO}
\end{equation}
where $\mathbf O_{WNOT}$ is the output from the trainable WNO model and $\mathbf O_{WNONT}$ is the output from the non-trainable WNO model. A $\beta$ value of unity is selected following the prevailing literature and the $\beta$ studies carried out in \ref{appendix beta studies}. A schematic for the proposed RP-WNO is shown in Fig. \ref{fig: RP-WNO}. 
To enable uncertainty quantification, we need to train multiple copies of RP-WNO, each with different initialization but the same architecture and training data. Since training any two copies of RP-WNO has no bearing on each other, we may train them in parallel, thus saving the overall computational time. Assuming $\mathcal P_i$ is the prediction from the $i^\text{th}$ copy of RP-WNO, the mean prediction $\mathcal P_M$ can be obtained as,
\begin{equation}
    \mathcal P_m = \dfrac{1}{n_c}\sum\limits_{i=1}^{n_c}\mathcal P_i,
    \label{eq: mean pred}
\end{equation}
where $n_c$ denotes the number of copies of RP-WNO trained for the same input data. The standard deviation $\mathcal P_s$ of the predictions can be obtained as follows,
\begin{equation}
    \mathcal P_s = \sqrt{\dfrac{\sum\limits_{i=1}^{n_c}(\mathcal P_i-\mathcal P_m)^2}{n_c}}.
    \label{eq: std pred}
\end{equation}
Note that if $\mathcal P_i$ is a vector or a multidimensional array, the operations stipulated in Eqs. \eqref{eq: mean pred} and \eqref{eq: std pred} will be implemented as element-wise operations. The $95\%$ confidence intervals can now be computed as $\mathcal P_m\pm1.96\mathcal P_s$. Algorithm \ref{algorithm: RP-WNO} details the steps involved in training the proposed RP-WNO ensemble. It should be noted that the outputs for the non-trainable network corresponding to the training data can be computed after the initialization step and stored for use later while training to save computational time. Additionally, multiple RP-WNO models in the RP-WNO ensemble can be trained simultaneously in parallel, as training for one has no bearing on the other.
\begin{algorithm}[ht!]
\caption{Training algorithm for the RP-WNO.}
\label{algorithm: RP-WNO}
\begin{algorithmic}[1]
\For{i = 1\,:\,no. of RP-WNO networks in the ensemble}
\State Initialize weights and bias for the trainable WNO and the non-trainable WNO with the \Statex\,\,\,\,\,\,\,\,\, same architecture.
\For{j = 1\,:\,no. of epochs}
\For{k = 1\,:\,no. of batches}
\State Input training data to trainable network and obtain $\mathcal O_{WNOT}$.        
\State Input training data to non-trainable network and obtain $\mathcal O_{WNONT}$.   \State Compute the combined output, $\mathcal O_{RP-WNO} = \mathcal O_{WNOT}+\mathcal O_{WNONT}$.
\State Compute the relevant loss function using $\mathcal O_{RP-WNO}$.
\State Perform backpropagation and update the trainable network's weights and biases.
\EndFor
\EndFor
\State Save the trained models for prediction later. 
\EndFor
\renewcommand{\algorithmicrequire}{\textbf{Output:}}
\Require{Ensemble of trained RP-WNO networks.}
\end{algorithmic}
\end{algorithm}
\section{Numerical illustrations}\label{section: Numerical Illustrations}
This section covers four case studies to illustrate the efficacy of the proposed RP-WNO. The first case study (CS-I) deals with a one-dimensional, time-dependent Burger's equation, while the second case study (CS-II) deals with a two-dimensional Darcy flow equation. The third case study (CS-III) aims to solve the seismic wave equation for variable velocity models, while the fourth case study (CS-IV) aims to link atomic structural defects to mesoscale properties in crystalline solids. The datasets for the first two case studies have been generated using the codes given in paper \cite{tripura2022wavelet,li2020fourier}. The velocity fields for the third case study have been pulled from \cite{deng2021openfwi}.
For the fourth case study, the datasets are pulled from \cite{yang2022linking}. The RP-WNO ensembles in the first and second case study utilize discrete wavelet transform, while the RP-WNO ensembles in the third and fourth case study utilize continuous wavelet transforms within the RP-WNO architecture. Further details regarding individual case studies are provided in the respective subsections.
\subsection{CS-I: Burger's equation}
The one-dimensional Burger's equation under consideration has periodic boundary conditions and is defined as,
\begin{equation}
    \begin{gathered}
    \dfrac{\partial u(x,t)}{\partial t}+u\dfrac{\partial u(x,t)}{\partial x} = \nu\dfrac{\partial^2 u(x,t)}{\partial x^2},\\
    x\in[0,1],\,\,\,\,\,t\in(0,1]\\
    u(x,0)\sim\mathcal N(0,625(-\Delta+25\,\mathbb I)^{-2})
    \end{gathered}
    \label{eqn: Burgers}
\end{equation}
where $\nu$ is the viscosity of the flow being modeled. For the current case study, $\nu$ is taken equal to 0.01. The initial condition for Burger's equation is sampled from a Gaussian random field as shown in Eq. \eqref{eqn: Burgers}, where $\Delta$ is a Laplacian operator. The mapping between the initial condition $u(x,0)$ and the velocity field obtained at $t = 1$, $u(x,1)$ is learned using the proposed RP-WNO.
\begin{figure}[ht!]
    \centering
    \includegraphics[width = \textwidth]{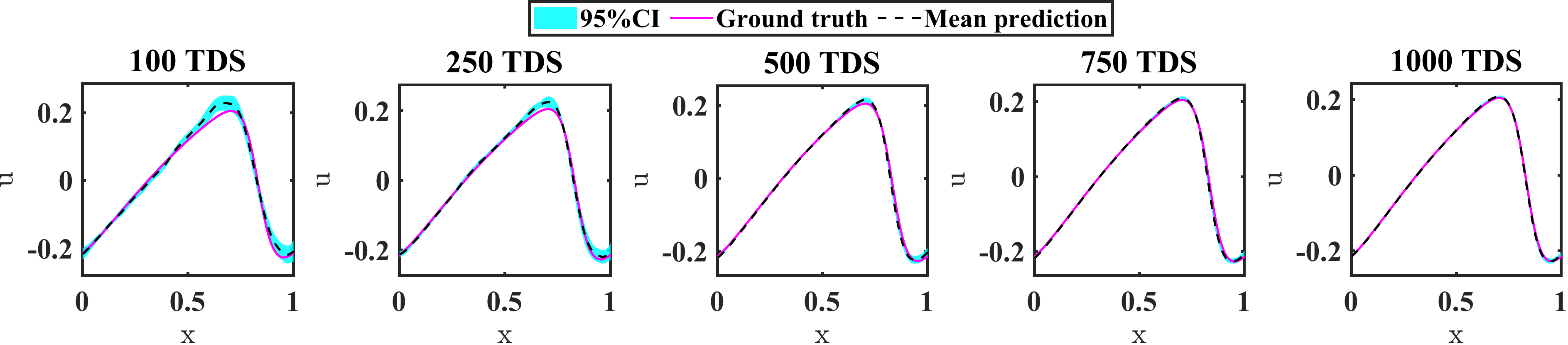}
    \caption{RP-WNO predictions when trained for different numbers of Training Data Samples (TDS) in CS-I.}
    \label{fig: Burger TDS}
\end{figure}
Fig. \ref{fig: Burger TDS} shows the mean prediction and confidence intervals obtained using the RP-WNO corresponding to a varying number of Training Data Samples (TDS). It is observed that the uncertainty associated with the prediction decrease as the data available for training increase, which is the expected behavior. Additionally, the mean prediction converges to the ground truth with an increase in TDS.
\begin{table}[ht!]
\centering
\caption{MAE and mean Std. of RP-WNO predictions in CS-I.}
\begin{tabular}{m{10em}m{3em}m{3em}m{3em}m{3em}m{3em}}
\hline
Training samples & 100 & 250 & 500 & 750 & 1000 \\\hline
MAE & 0.0192 & 0.0098 & 0.0046 &  0.0033 &  0.0025\\
Mean Std. & 0.0107 & 0.0072 & 0.0039 & 0.0031 & 0.0027 \\
\hline
\end{tabular}
\label{table: Burger TDS}
\end{table}
Table \ref{table: Burger TDS} quantifies the results shown in Fig. \ref{fig: Burger TDS}. To generate the Mean Absolute Error (MAE) and mean Standard deviation (Std.), a prediction ensemble with 8000 samples was generated using the trained RP-WNO ensemble. To calculate MAE values, mean predictions from RP-WNO were compared against the ground truth, and for mean Std. values, the variation observed within the RP-WNO ensemble was considered. As can be seen that the trend observed in Table \ref{table: Burger TDS}, further solidifies the observations associated with Fig. \ref{fig: Burger TDS}.

\begin{figure}[ht!]
\centering
\begin{subfigure}{0.40\textwidth}
    \centering
    \includegraphics[width = \textwidth]{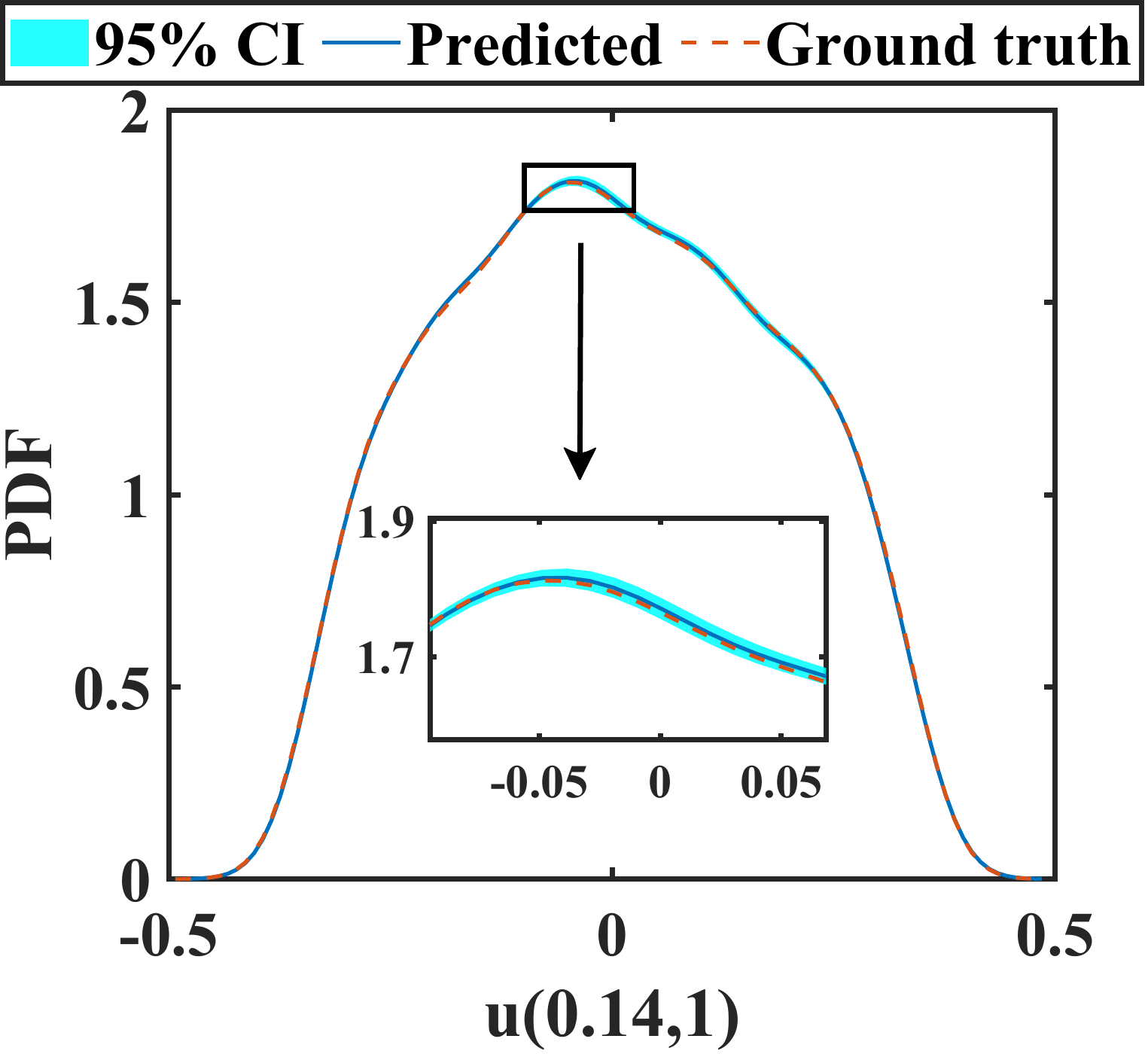}
    \caption{PDF computed at $x = 0.14$ and $t = 1$.}
\end{subfigure}
\begin{subfigure}{0.40\textwidth}
    \centering
    \includegraphics[width = \textwidth]{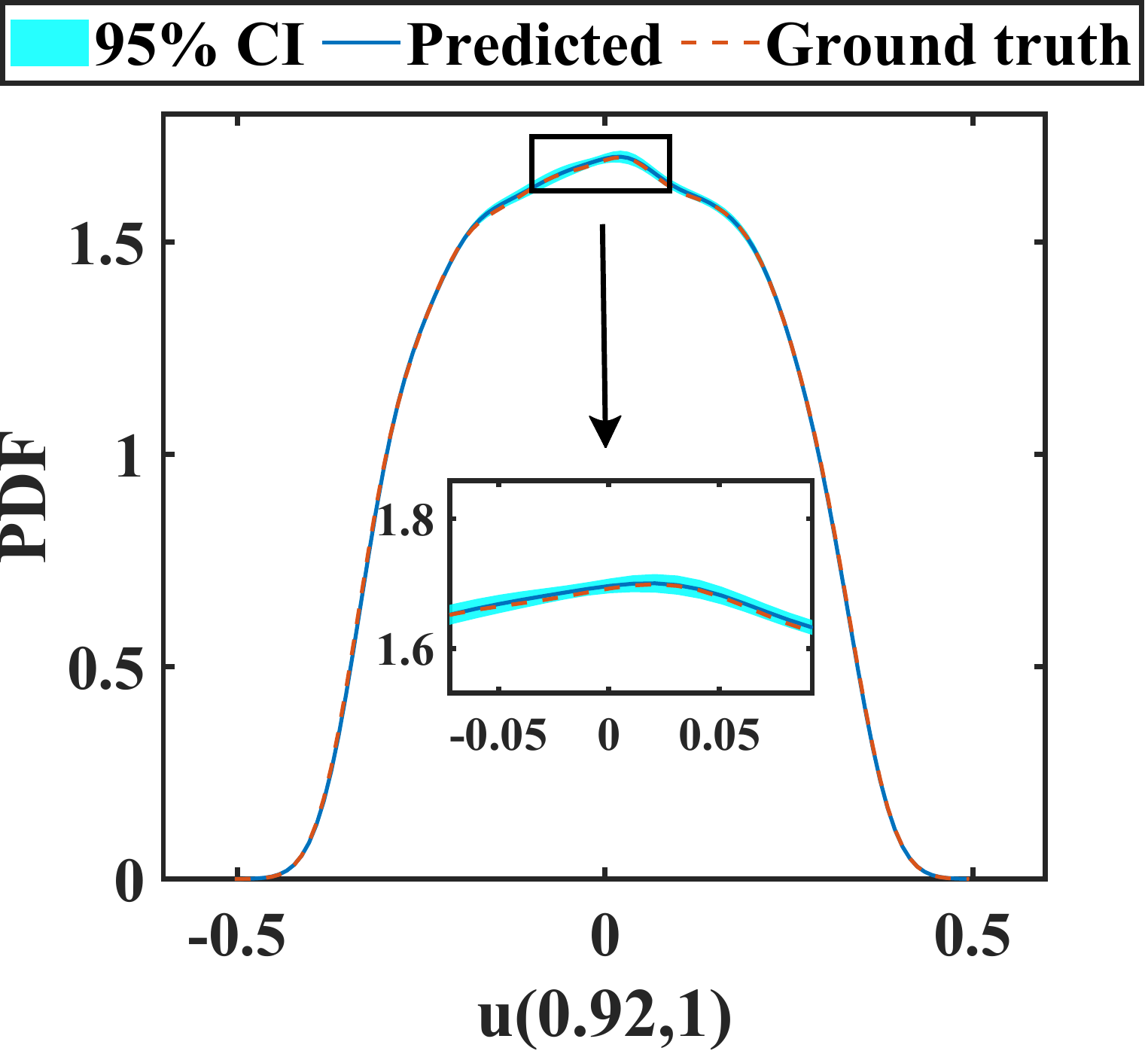}
    \caption{PDF computed at $x = 0.92$ and $t = 1$.}
\end{subfigure}
\begin{subfigure}{0.40\textwidth}
    \centering
    \includegraphics[width = \textwidth]{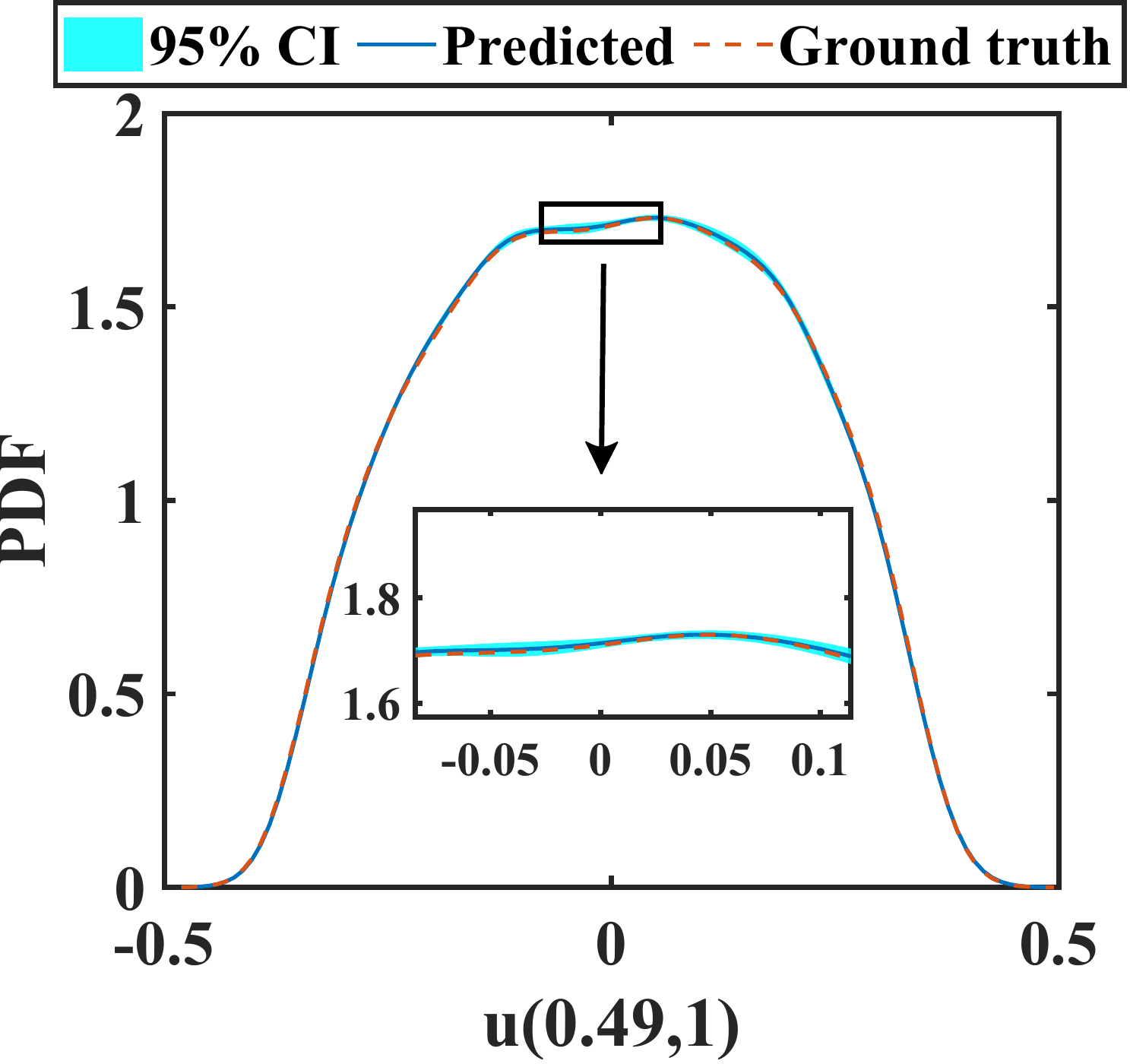}
    \caption{PDF computed at $x = 0.49$ and $t = 1$.}
\end{subfigure}
    \caption{Alleatoric uncertainty estimates for CS-I.}
    \label{fig: Burger aleatoric}
\end{figure}
Fig. \ref{fig: Burger aleatoric} shows the aleatoric uncertainty estimates compared against the ground truth for the case when the RP-WNO model is trained for 1000 training samples. RP-WNO predictions closely follow the ground truth and provide additional confidence intervals that effectively capture the ground truth.
\subsection{CS-II: Darcy flow equation}
The time-independent two-dimensional Darcy flow equation is given as,
\begin{equation}
\begin{gathered}
    -\nabla(a(x,y)\nabla u(x,y)) = 1,\\
    x\in(0,1),\,\,\,\,\,y\in(0,1),\\
    a(x,y)\sim\psi\mathcal N(0,(-\Delta+9\,\mathbb I)^{-2}),
    \end{gathered}
    \label{eqn: Darcy}
\end{equation}
where $a(x,y)$ is the permeability field and is sampled from the Gaussian random field $\mathcal N(0,(-\Delta+9\,\mathbb I)^{-2})$. $\Delta$ in Eq. \eqref{eqn: Darcy} is Laplacian with zero Neumann boundary condition, and $\psi$ is a mapping that transforms the function such that the positive values are set to 12 and the negative values to 3. Mapping for this case study is carried out between the permeability field $a(x,y)$ and the corresponding pressure $u(x,y)$.
\begin{figure}[ht!]
    \centering
    \includegraphics[width = \textwidth]{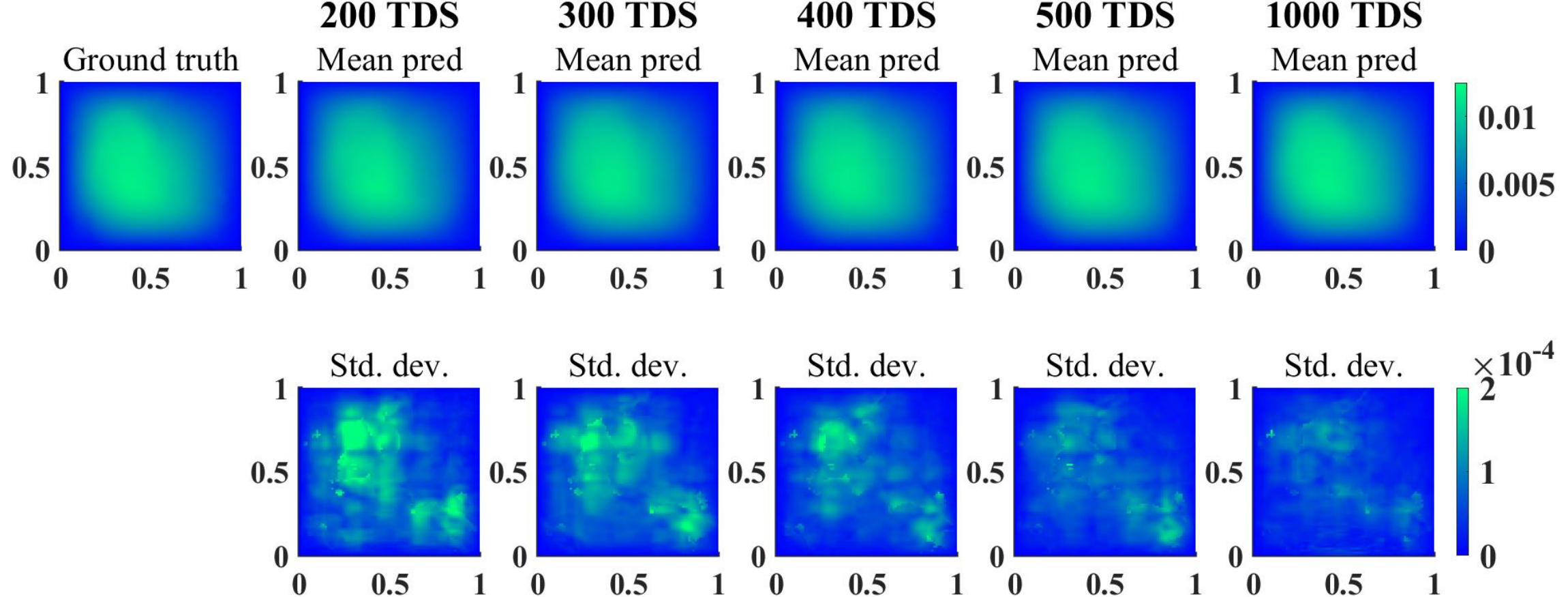}
    \caption{ RP-WNO predictions when trained for different numbers of Training Data Samples (TDS) in CS-II.}
    \label{fig: Darcy TDS}
\end{figure}
Similar to the previous case study, Fig. \ref{fig: Darcy TDS} shows the prediction results for the two-dimensional Darcy flow case study when the RP-WNO ensemble is trained for different numbers of training samples. The results follow the expected trend, and the variation observed in prediction reduces with an increase in the training data.
\begin{table}[ht!]
\centering
\caption{MAE and mean Std. of RP-WNO predictions in CS-II.}
\begin{tabular}{m{10em}m{3em}m{3em}m{3em}m{3em}m{3em}}
\hline
Training samples & 200 & 300 & 400 & 500 & 1000\\\hline
MAE ($\times10^{-3}$) & 0.1280 & 0.0962 & 0.0770 &  0.0631 & 0.0408\\
Mean Std. ($\times10^{-4}$) & 0.7224 & 0.6025 & 0.5323 & 0.4813 & 0.3601\\
\hline
\end{tabular}
\label{table: Darcy TDS}
\end{table}
The trend for the evolution of MAE and mean Std. with the increase in training data is shown in Table \ref{table: Darcy TDS}. Eigth thousand test samples were used to arrive at the presented results. Both MAE and mean Std. decrease monotonically, which was the expected behavior.

\begin{figure}[ht!]
\centering
\begin{subfigure}{0.40\textwidth}
    \centering
    \includegraphics[width = \textwidth]{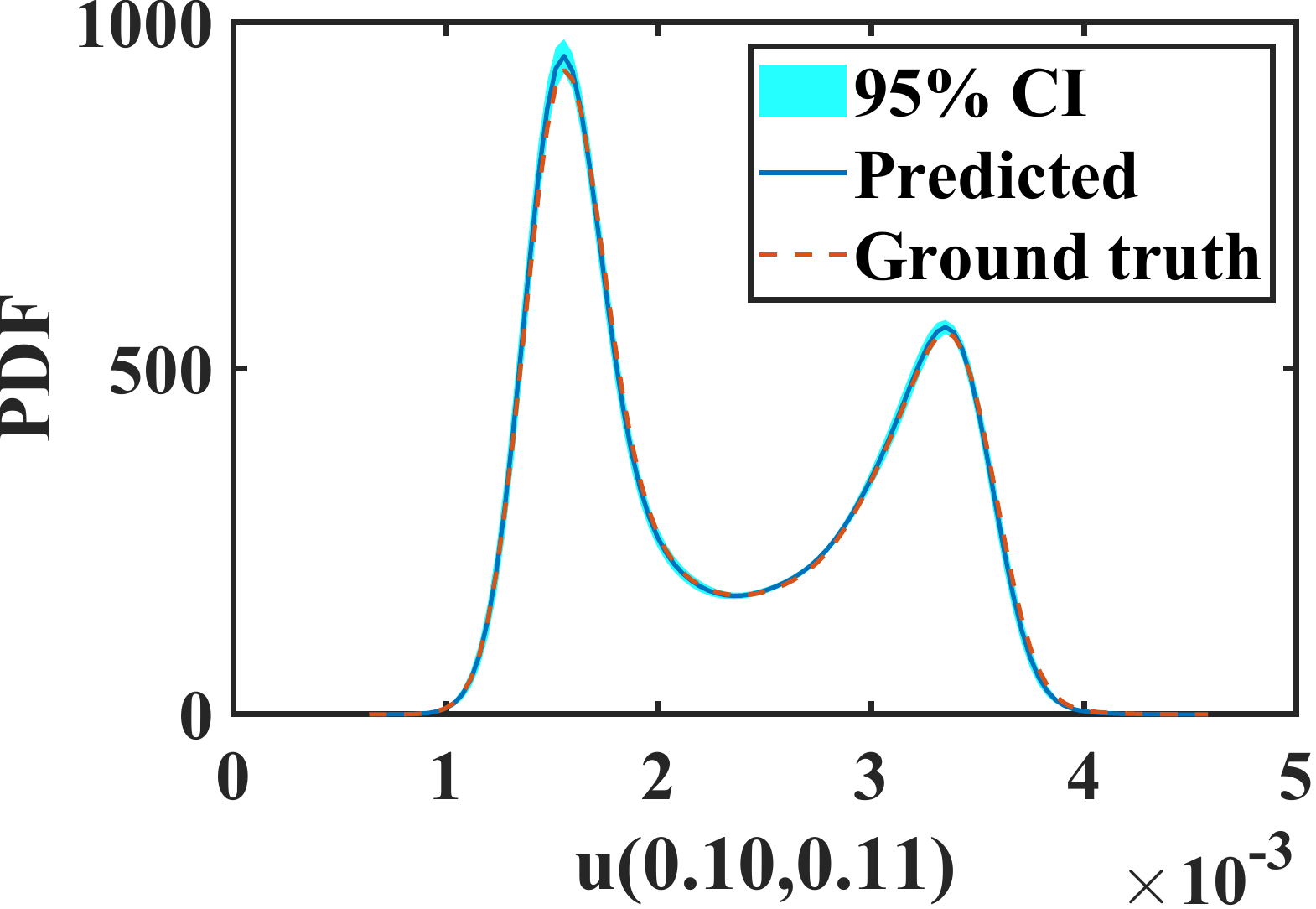}
    \caption{PDF computed at $x = 0.10$ and $y = 0.11$.}
\end{subfigure}
\begin{subfigure}{0.40\textwidth}
    \centering
    \includegraphics[width = \textwidth]{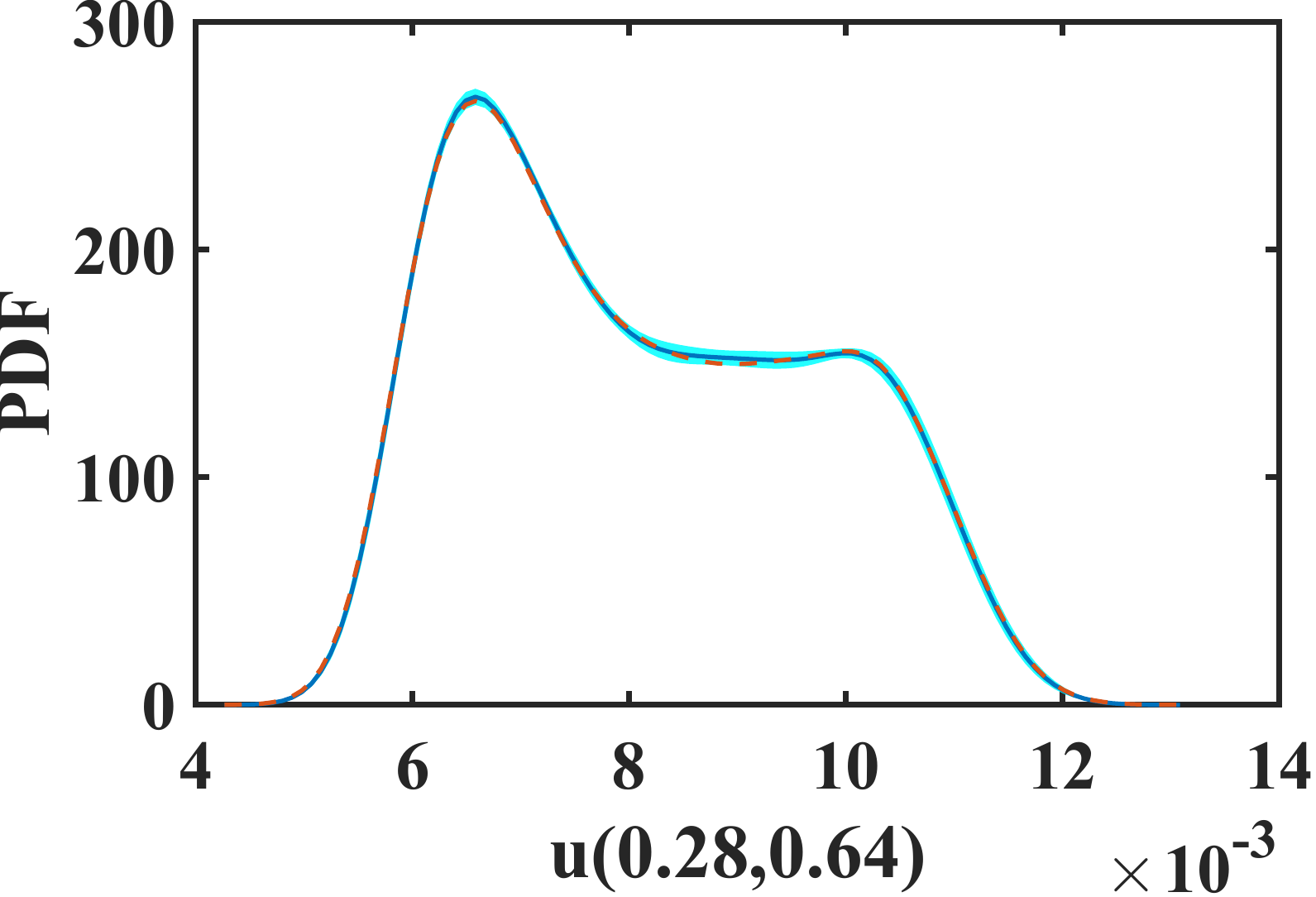}
    \caption{PDF computed at $x = 0.28$ and $y = 0.64$.}
\end{subfigure}
\begin{subfigure}{0.40\textwidth}
    \centering
    \includegraphics[width = \textwidth]{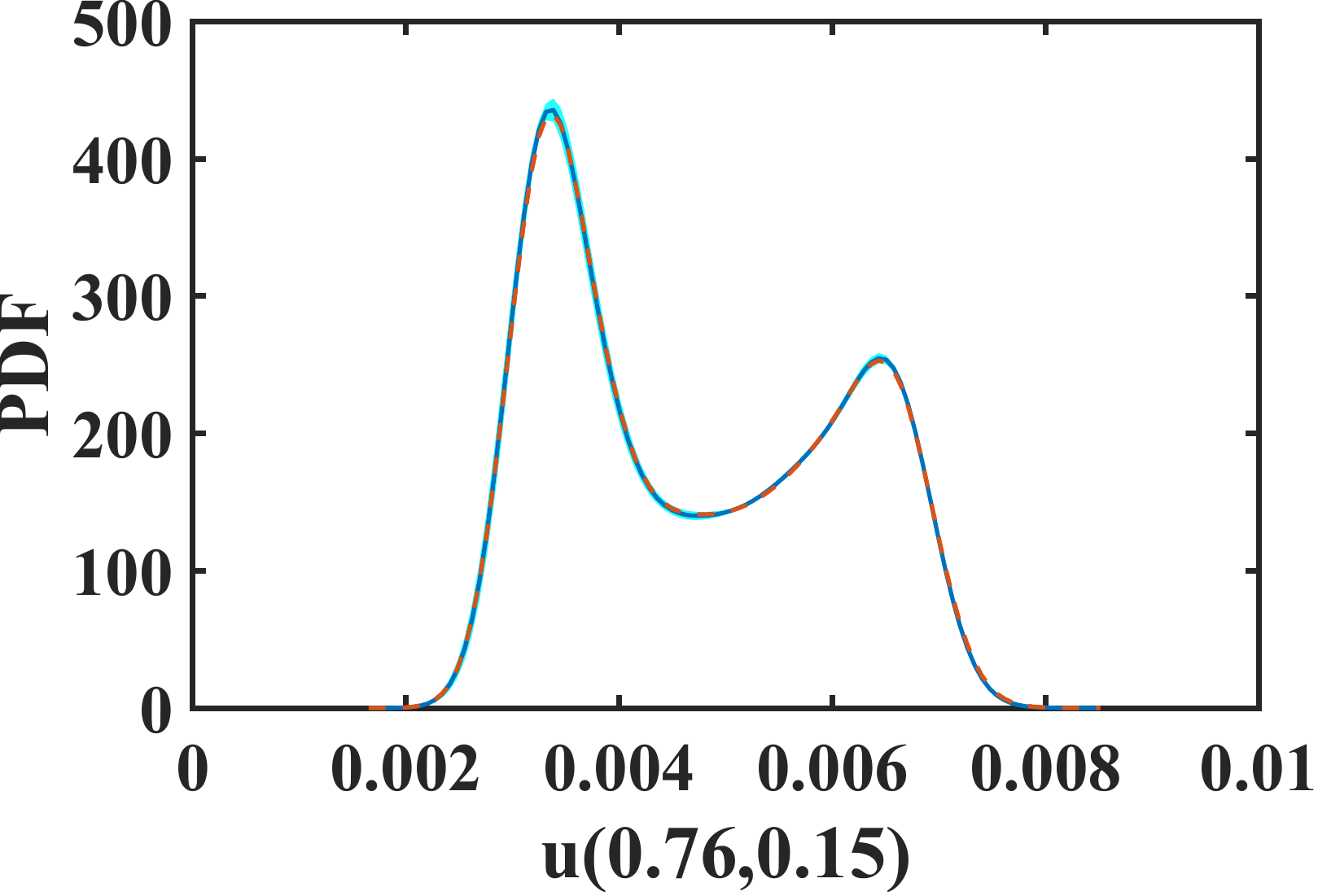}
    \caption{PDF computed at $x = 0.76$ and $y = 0.15$.}
\end{subfigure}
\begin{subfigure}{0.40\textwidth}
    \centering
    \includegraphics[width = \textwidth]{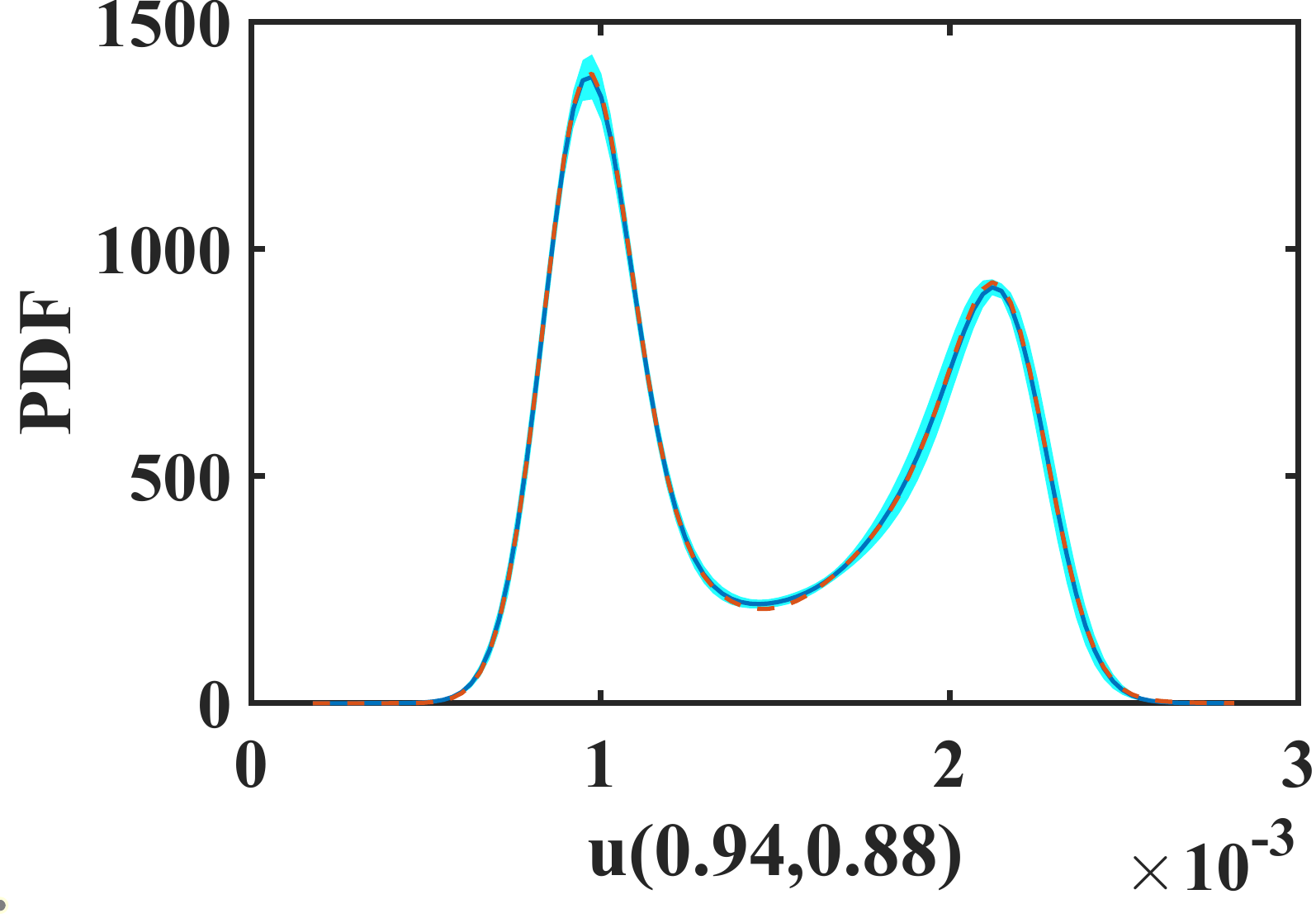}
    \caption{PDF computed at $x = 0.94$ and $y = 0.88$.}
\end{subfigure}
    \caption{Aleatoric uncertainty estimated for two-dimensional Darcy flow in CS-II.}
    \label{fig: Darcy alleatoric}
\end{figure}
Fig. \ref{fig: Darcy alleatoric} shows the aleatoric uncertainty estimates for the two-dimensional Darcy flow equation in CS-II. As can be seen, the mean prediction closely follows the ground truth, and we also obtain additional confidence intervals associated with the proposed RP-WNO predictions. This clearly illustrates the proposed RP-WNO can capture such multimodal distributions.  The RP-WNO ensemble was trained using 1000 training samples to obtain the results.
\subsection{CS-III: Helmholtz equation}
For the third case study, we will consider the Helmholtz equation, the solution for which gives pressure wavefields in the frequency domain. Its use case can be found in applications where the effect of traveling waves has to be studied, for example, in electromagnetic radiation studies, seismic ground exploration, etc. The Helmholtz equation for waves traveling in two dimensions is defined as,
\begin{equation}
    \nabla^2 u(x,z;w) + k^2u(x,z;w)=f(x_s,z_s;w),
\end{equation}
where $u$ is the pressure wavefield in the frequency domain, and $w$ is the angular frequency. $k$ is the wave number, and the point source is located at the location $(x_s,z_s)$. In the current case study, for dataset generation, the Helmholtz equation is not solved directly, but instead, the two-dimensional acoustic wave equation is solved first in the time domain, and the results produced are then transferred to the frequency domain in order to obtain the pressure wavefield $u(x,z)$. The two-dimensional acoustic wave equation is defined as,
\begin{equation}
    \nabla^2 p(x,z,t) - \dfrac{1}{v(x,z)^2}\dfrac{\partial^2p(x,z,t)}{\partial t^2}=s(x_s,z_s,t),
    \label{eq: acoustic wave eqn}
\end{equation}
where $p(x,z,t)$ is the pressure wavefield in time domain and $v(x,z)$ is the velocity of wave at location $(x,z)$. The point source $s$ is located at the location $(x_s,z_s)$. The spatial domain $(x,z)\in[0\,m,690\,m]^2$ in Eq. \eqref{eq: acoustic wave eqn} is discretized with $\Delta x = \Delta z = 10\,m$ and a constant source of magnitude $1$ is selected for modeling. The variability in different samples is introduced by virtue of changing velocity models, which are obtained from \cite{deng2021openfwi} (FlatVel-A dataset, Vel Family). The source location is fixed at $x_s = 680\,m$ and $z_s = 10\,m$. Now, the pressure wavefield in the frequency domain can be computed for a wide range of frequencies, but the initial lower frequencies will contribute the majority of the energy. As a result, to test the efficacy of the proposed RP-WNO, the dataset created contains pressure wavefields corresponding to frequencies 1Hz, 3Hz, 5Hz, 7Hz, and 9Hz.  

Now, although we can use different RP-WNO ensembles for mapping velocity field to pressure wavefield of different frequencies, to show the efficacy of the proposed operator learning framework, the mapping in the current case study is carried out for all five frequencies using a single RP-WNO ensemble. To achieve this, the input to RP-WNO, which was supposed to be the velocity field and the grid on which it is discretized, is concatenated with an additional matrix that contains the frequency information corresponding to which output is obtained.
The additional matrix being concatenated has the same dimensions as the velocity wavefield, and its elements have a constant value equal to the frequency under consideration. For example, if corresponding to velocity field $v(x,z) = \mathbf V_1$, we have pressure wavefield $u(x,z;w) = \mathbf U_1^{(w)}$, we will obtain five samples for training the proposed RP-WNO, which can be visualized as follows,
\begin{equation}
    \begin{matrix}
    \textbf{Inputs} & \textbf{Outputs}\\
    \mathbf V_1,\,1\,\mathcal U,\,\mathcal G_x,\,\mathcal G_z & \mathbf U_1^{(1)}\\
    \mathbf V_1,\,3\,\mathcal U,\,\mathcal G_x,\,\mathcal G_z & \mathbf U_1^{(3)}\\
    \mathbf V_1,\,5\,\mathcal U,\,\mathcal G_x,\,\mathcal G_z & \mathbf U_1^{(5)}\\
    \mathbf V_1,\,7\,\mathcal U,\,\mathcal G_x,\,\mathcal G_z & \mathbf U_1^{(7)}\\
    \mathbf V_1,\,9\,\mathcal U,\,\mathcal G_x,\,\mathcal G_z & \mathbf U_1^{(9)},
    \end{matrix}
\end{equation}
where $\mathcal U$ is a unity matrix with the same dimensions as the matrix $\mathbf V_1$. $\mathcal G_x$ and $\mathcal G_z$ are the matrices containing the grid information on which $\mathbf V_1$ is discretized. To train the RP-WNO ensembles, 500 velocity models are used. Additionally, the pressure wavefield in the frequency domain will be represented by complex numbers. Hence, the real and imaginary components of the pressure wavefield $u(x,z;w)$ are trained separately using two different RP-WNO ensembles.
\begin{figure}[ht!]
    \centering
    \includegraphics[width = 0.70\textwidth]{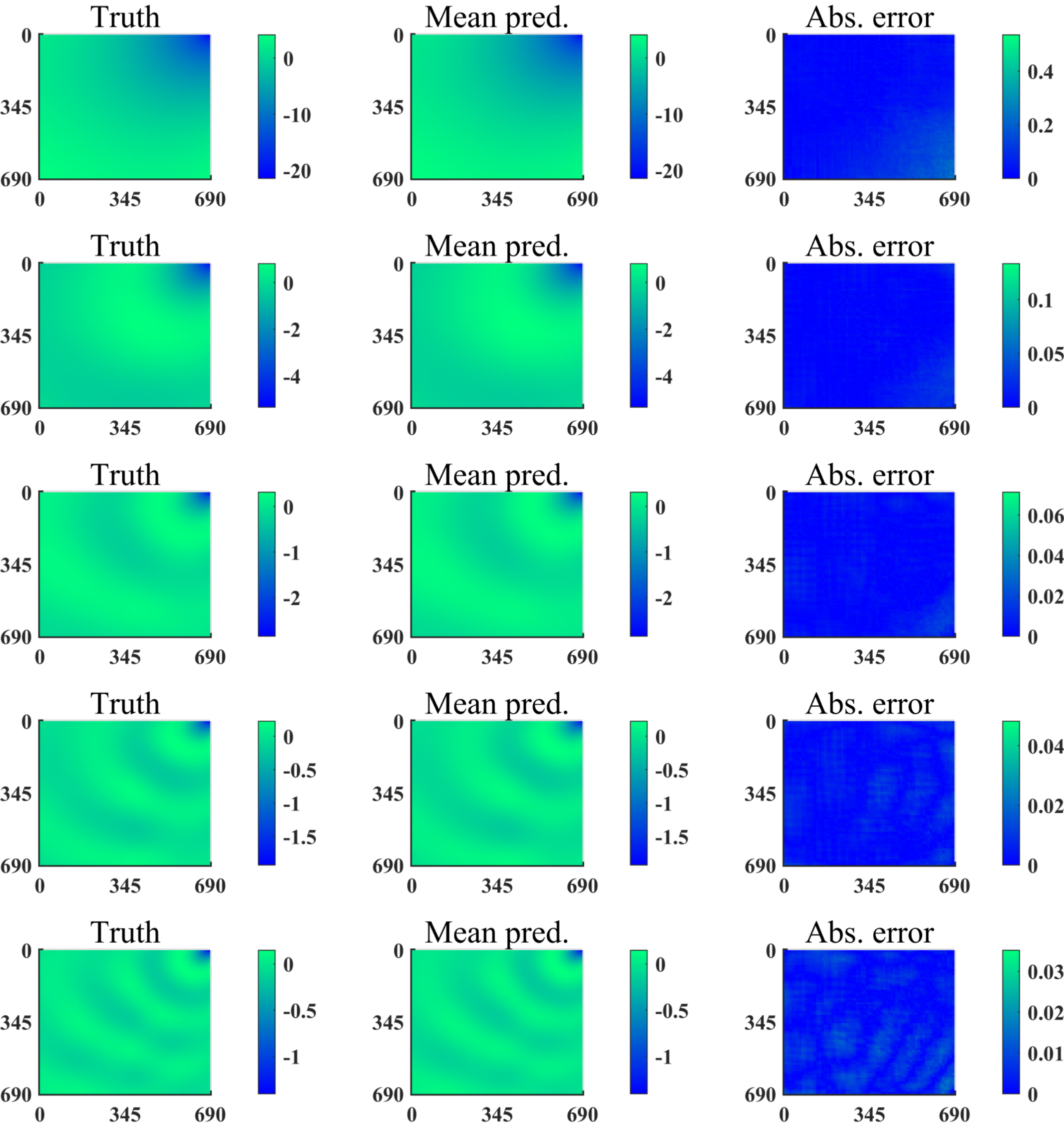}
    \caption{Pressure wavefield predictions (real part) corresponding to one sample of the velocity field, obtained using the trained RP-WNO ensemble. The result corresponding to a frequency equal to 1Hz is given at the top, and the one corresponding to 9Hz is given at the bottom. Results in between correspond to frequencies of 3Hz, 5Hz, and 7Hz, respectively, from top to bottom.}
    \label{fig: CSIII real part}
\end{figure}
Fig. \ref{fig: CSIII real part} shows the mean RP-WNO predictions compared against the ground truth for the real part of the pressure wavefield. The results shown correspond to a frequency of 1Hz, 3Hz, 5Hz, 7Hz, and 9Hz, from top to bottom, and as can be seen, the mean prediction from RP-WNO closely follows the ground truth.

\begin{figure}[ht!]
\centering
\begin{subfigure}{0.475\textwidth}
    \centering
    \includegraphics[width = \textwidth]{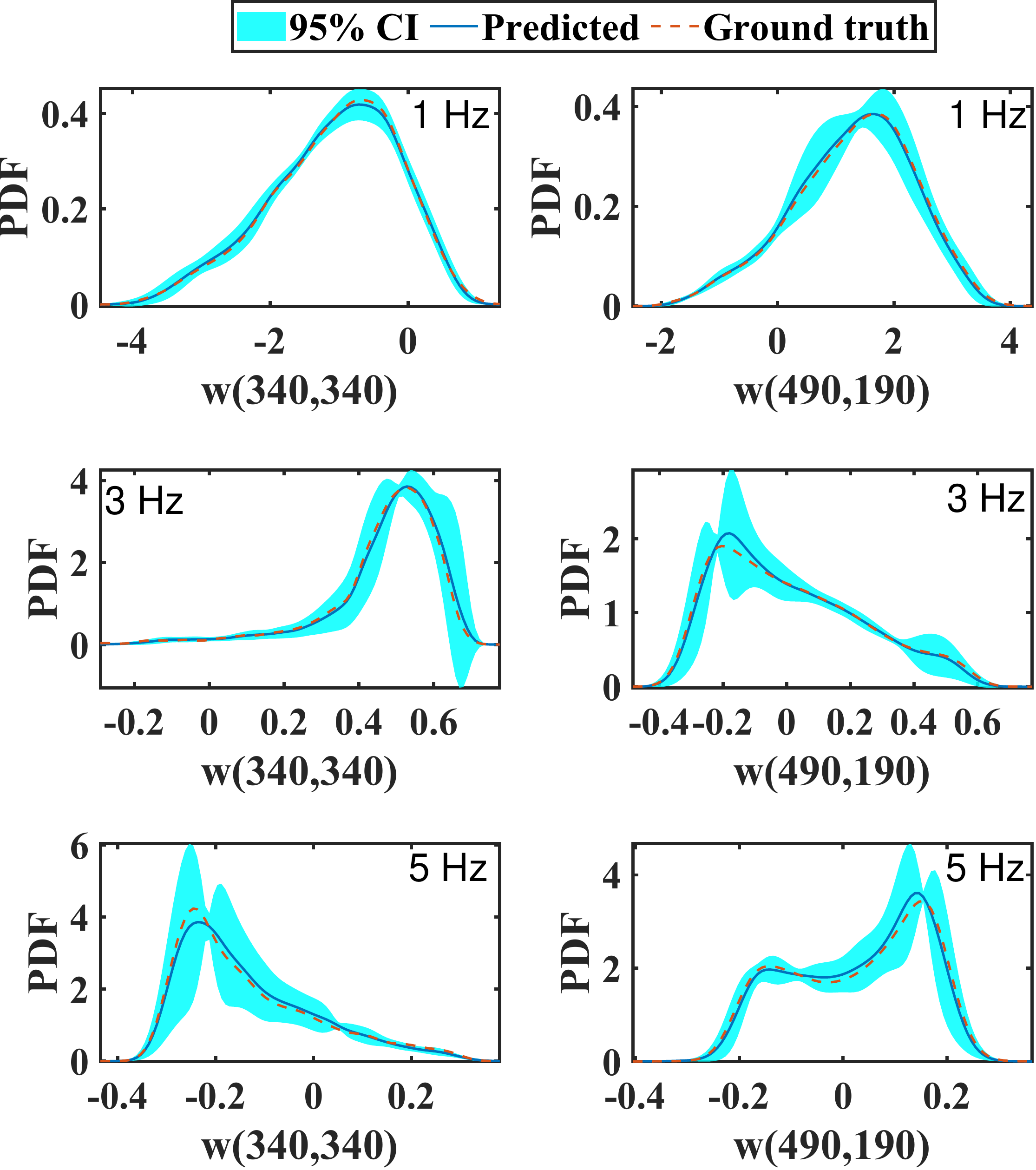}
    \caption{Discrete wavelet transform}
\end{subfigure}
\begin{subfigure}{0.475\textwidth}
    \centering
    \includegraphics[width = \textwidth]{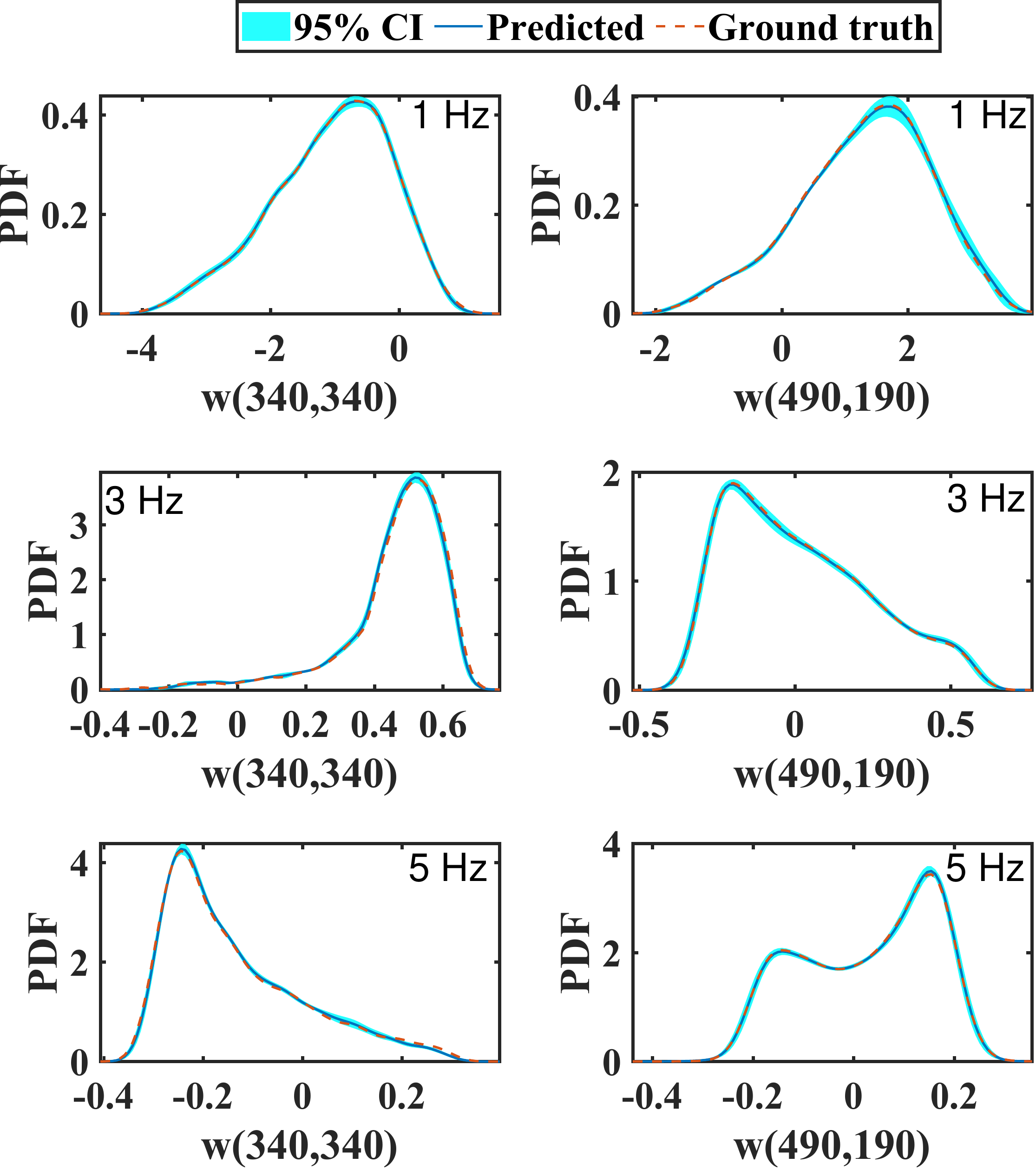}
    \caption{Continuous wavelet transform}
\end{subfigure}
    \caption{Alleatoric uncertainty estimates for the Pressure wavefield (real part), computed corresponding to different frequencies (1Hz, 3Hz, and 5Hz). PDFs are computed at two locations, (i)$x = 340$, $z = 340$ and (ii)$x = 490$, $z = 190$.}
    \label{fig: aleatoric uncertainty 1 Hz, 3 Hz, 5 Hz}
\end{figure}
\begin{figure}[ht!]
\centering
\begin{subfigure}{0.475\textwidth}
    \centering
    \includegraphics[width = \textwidth]{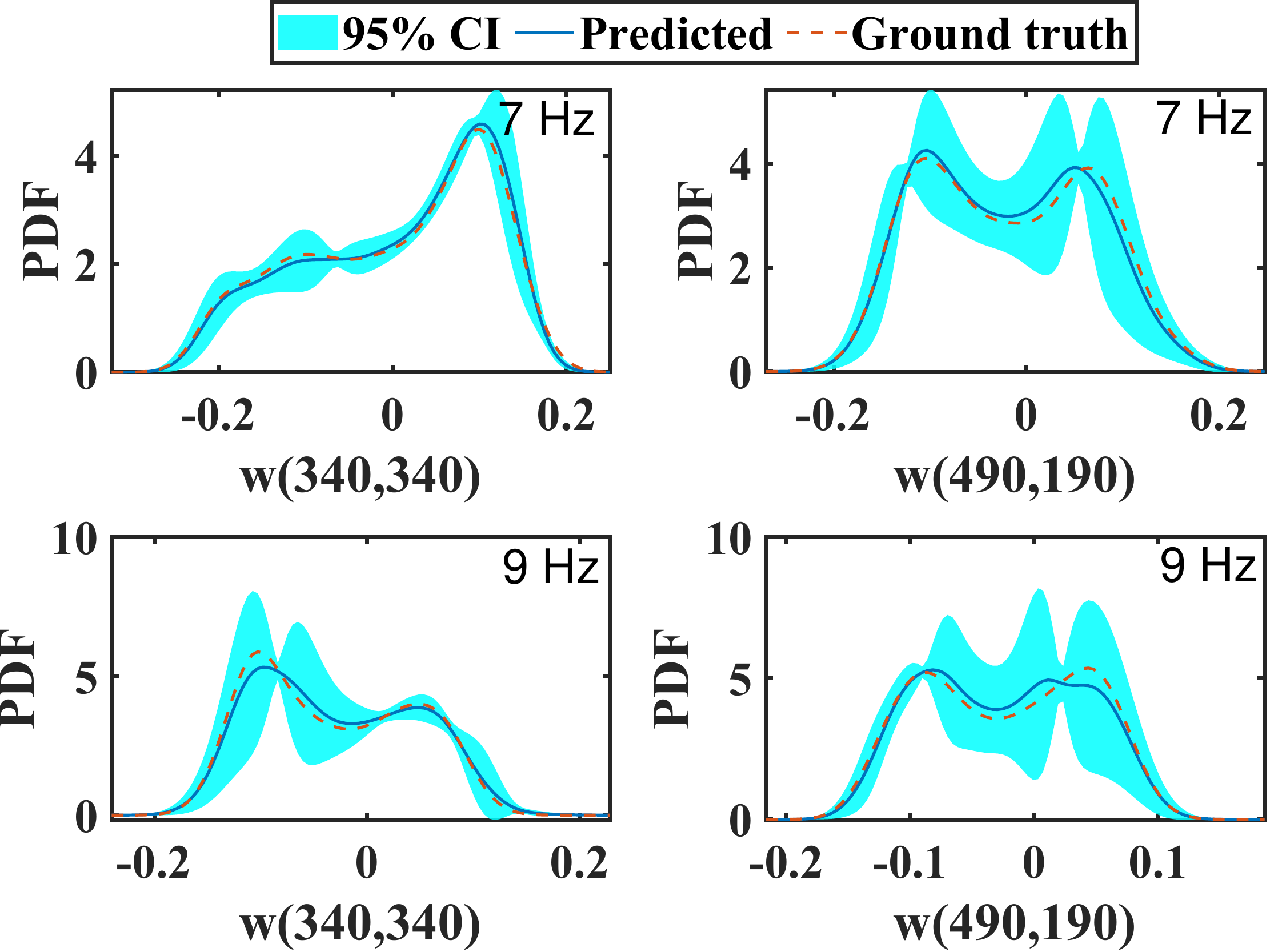}
    \caption{Discrete wavelet transform}
\end{subfigure}
\begin{subfigure}{0.475\textwidth}
    \centering
    \includegraphics[width = \textwidth]{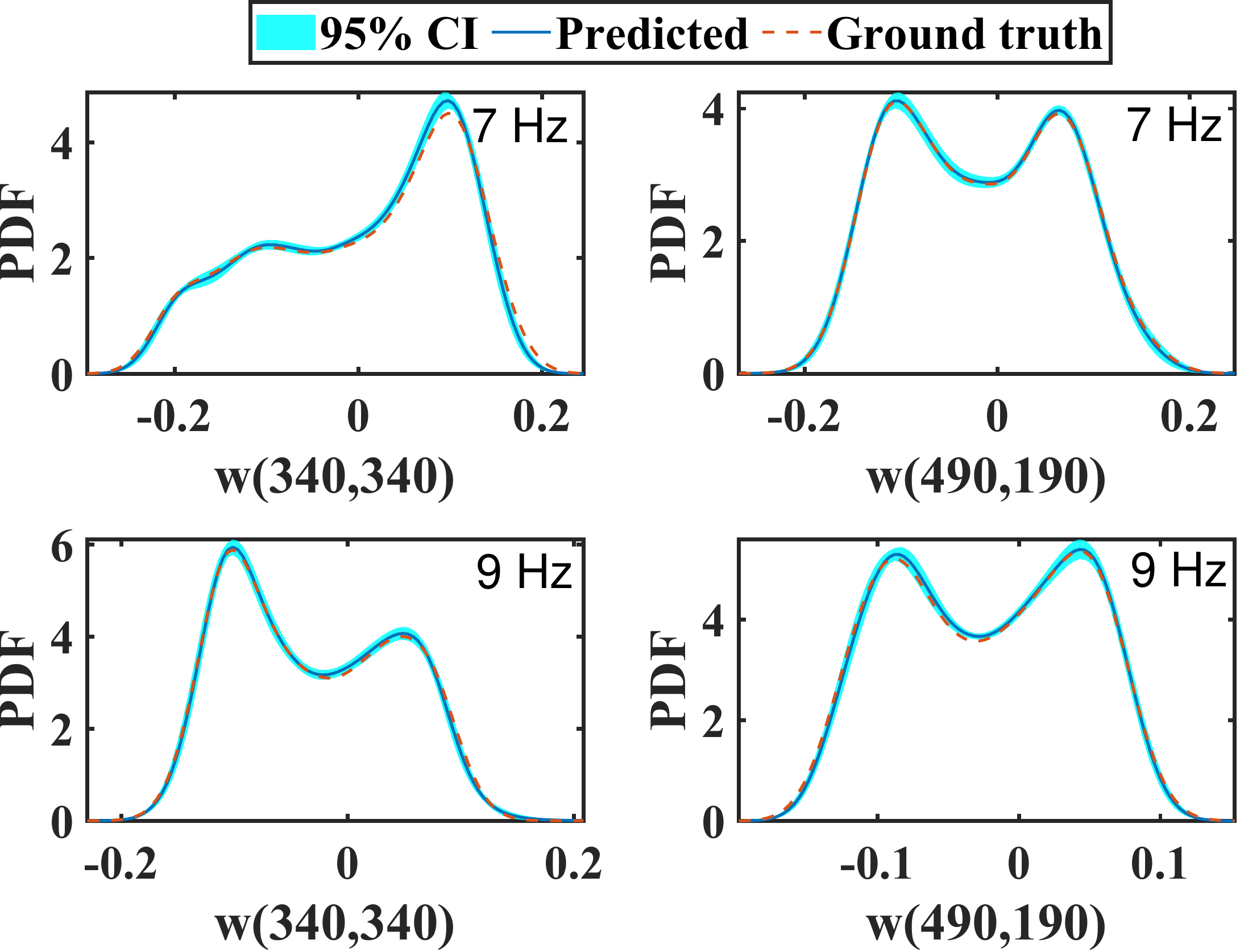}
    \caption{Continuous wavelet transform}
\end{subfigure}
    \caption{Alleatoric uncertainty estimates for the Pressure wavefield (real part), computed corresponding to different frequencies (7Hz and 9Hz). PDFs are computed at two locations, (i)$x = 340$, $z = 340$ and (ii)$x = 490$, $z = 190$.}
    \label{fig: aleatoric uncertainty 7 Hz, 9 Hz}
\end{figure}
Figs. \ref{fig: aleatoric uncertainty 1 Hz, 3 Hz, 5 Hz} and \ref{fig: aleatoric uncertainty 7 Hz, 9 Hz} show the aleatoric uncertainty estimates for the real part of the pressure wavefield corresponding to different frequencies. Five thousand different velocity models were used in order to obtain the stipulated results. Part (a) of Figs. \ref{fig: aleatoric uncertainty 1 Hz, 3 Hz, 5 Hz} and \ref{fig: aleatoric uncertainty 7 Hz, 9 Hz} show results corresponding to RP-WNO ensemble utilizing discrete wavelet transform. In this, among the four wavelet coefficients of the discrete wavelet transform, only the approximate and the horizontal components are learned while training. Although the mean predictions for the RP-WNO model trained using discrete wavelet transform, give a reasonable approximation of the ground truth, the uncertainty observed in predictions is huge owing to the slight deviations observed in the prediction results.

On the other hand, part (b) of Figs. \ref{fig: aleatoric uncertainty 1 Hz, 3 Hz, 5 Hz} and \ref{fig: aleatoric uncertainty 7 Hz, 9 Hz} show the predictions of the RP-WNO ensemble utilizing continuous wavelet transform. In this, all the wavelet coefficients of the continuous wavelet transform were learned, and as can be seen that the mean predictions in this case closely follow the ground truth, and correspondingly tight uncertainty bounds are obtained for the predicted results.

\begin{figure}[ht!]
    \centering
    \includegraphics[width = 0.75\textwidth]{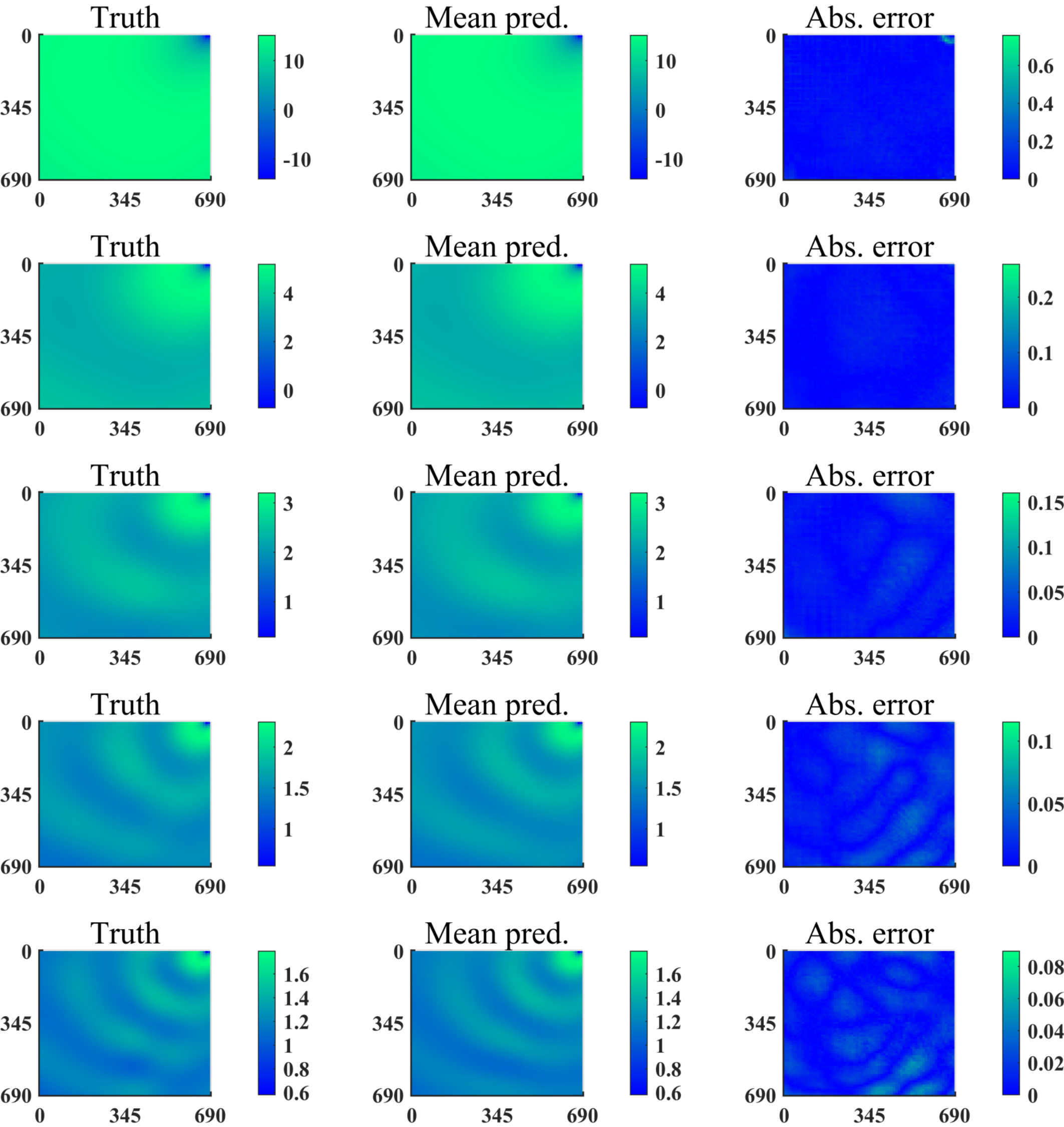}
    \caption{Pressure wavefield predictions (imaginary part) corresponding to one sample of the velocity field, obtained using the trained RP-WNO ensemble. The result corresponding to a frequency equal to 1Hz is given at the top, and the one corresponding to 9Hz is given at the bottom. Results in between correspond to frequencies of 3Hz, 5Hz, and 7Hz, respectively, from top to bottom.}
    \label{fig: CSIII imaginary part}
\end{figure}
Fig. \ref{fig: CSIII imaginary part} shows the mean RP-WNO predictions compared against the ground truth for the imaginary part of the pressure wavefield. The results shown correspond to a frequency of 1Hz, 3Hz, 5Hz, 7Hz, and 9Hz, from top to bottom, and as can be seen, the mean prediction obtained from the trained RP-WNO ensemble, closely follows the ground truth.

\begin{figure}[ht!]
    \centering
    \includegraphics[width = 0.65\textwidth]{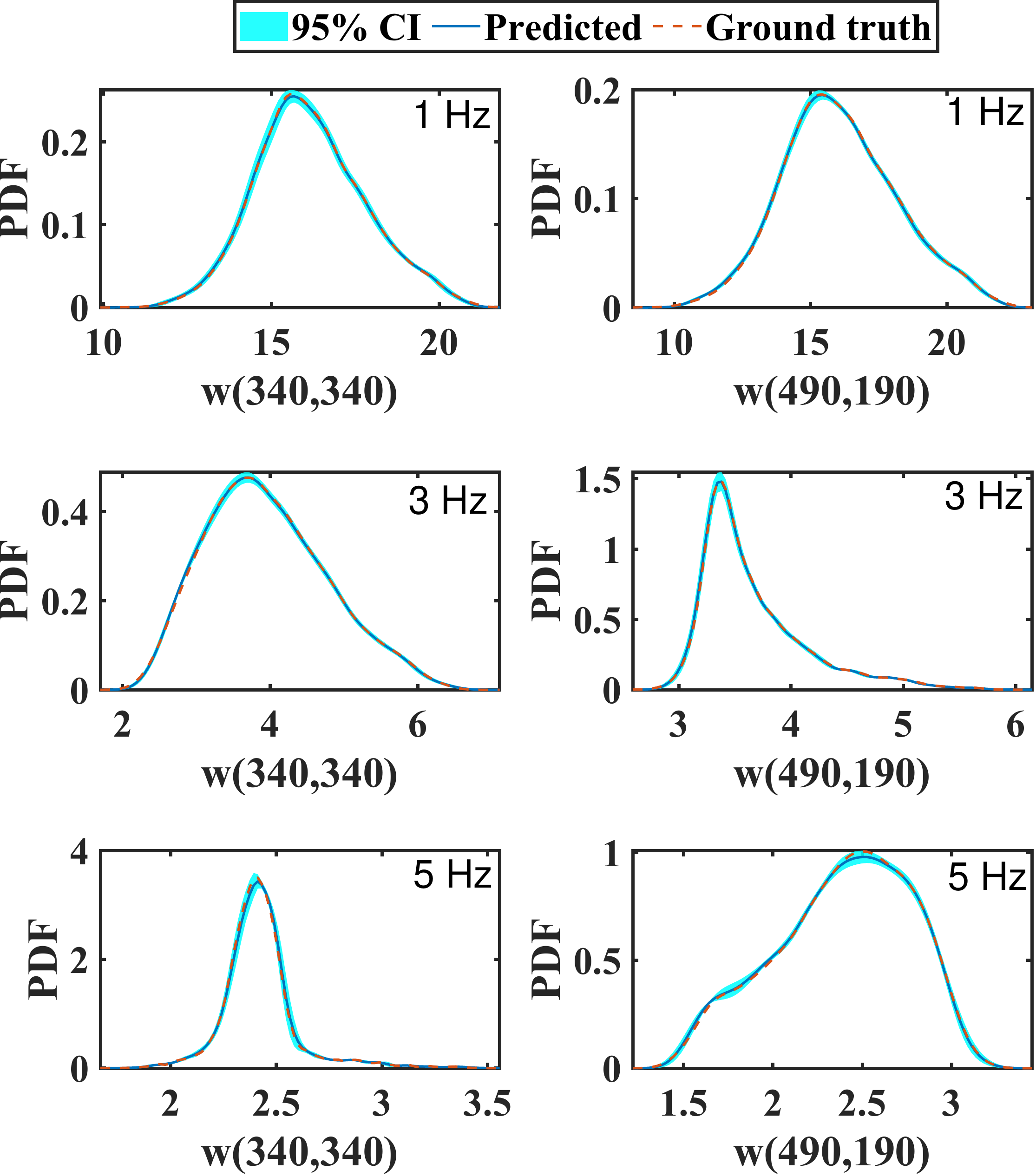}
    \caption{Alleatoric uncertainty estimates for the Pressure wavefield (imaginary part), computed corresponding to different frequencies (1Hz, 3Hz, and 5Hz).}
    \label{fig: imaginary aleatoric uncertainty 1 Hz, 3 Hz, 5 Hz}
\end{figure}
\begin{figure}[ht!]
    \centering
    \includegraphics[width = 0.65\textwidth]{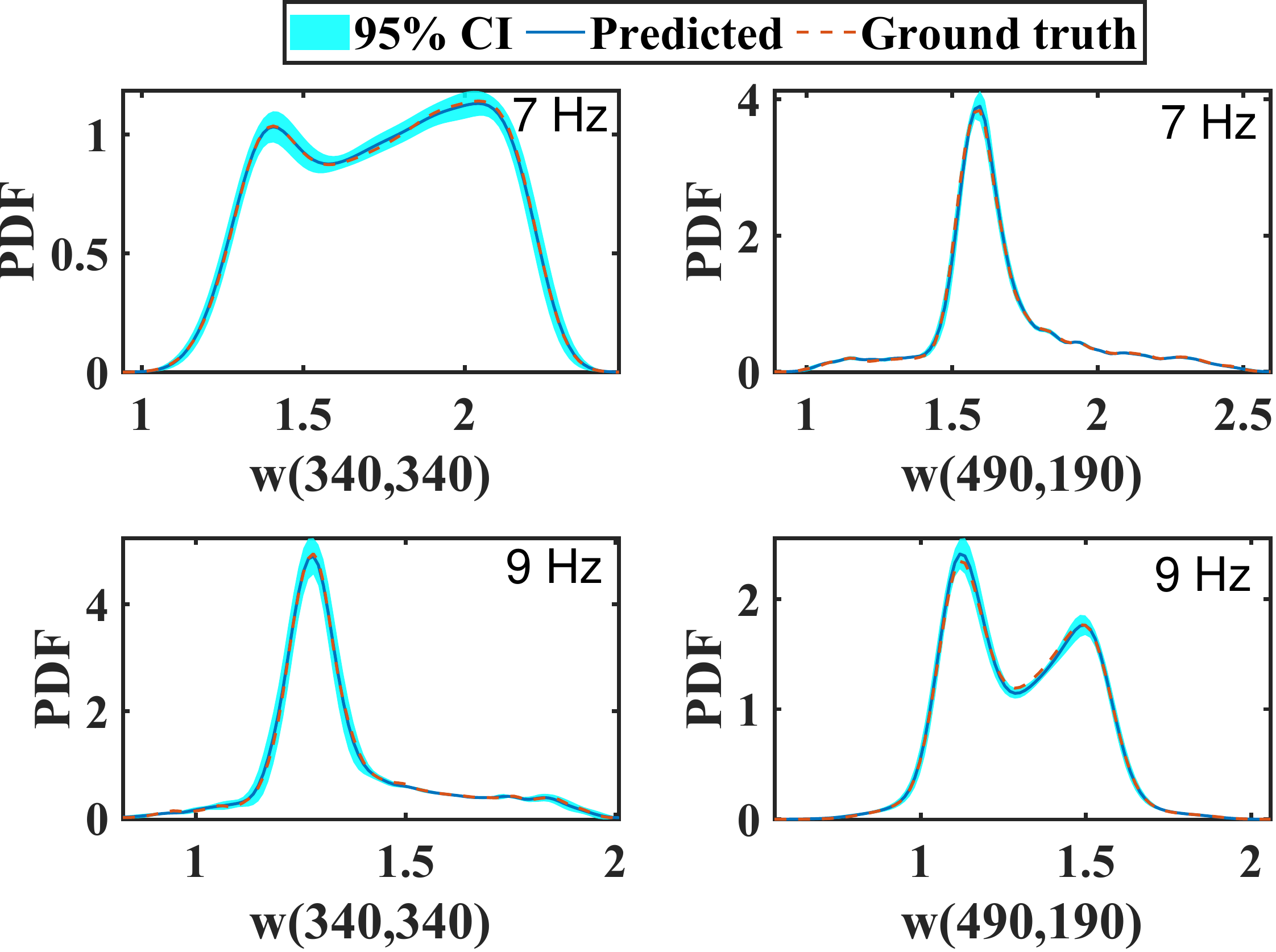}
    \caption{Alleatoric uncertainty estimates for the Pressure wavefield (imaginary part), computed corresponding to different frequencies (7Hz and 9Hz).}
    \label{fig: imaginary aleatoric uncertainty 7 Hz, 9 Hz}
\end{figure}
Figs. \ref{fig: imaginary aleatoric uncertainty 1 Hz, 3 Hz, 5 Hz} and \ref{fig: imaginary aleatoric uncertainty 7 Hz, 9 Hz} show the aleatoric uncertainty estimates for the real part of pressure wavefield corresponding to different frequencies. Five thousand different velocity models were used in order to obtain the stipulated results. The predictions are obtained using the RP-WNO model utilizing continuous wavelet transform. All the wavelet coefficients of the continuous wavelet transform were learned, and similar to the aleatoric uncertainty estimates of the real part, the mean predictions closely follow the ground truth giving a tight associated uncertainty bound.
\subsection{CS-IV: Mesoscale properties}
Having looked at some arguably conventional operator learning problems, the fourth case study will look to map the atomic structural defects to the mesoscale properties in crystalline solids. Structural defects are unavoidable when it comes to synthesizing materials, and the same affects the material properties like conductivity, elastic modulus, etc. Estimating the material property directly based on the atomic structure is non-trivial and requires extensive simulations and experiments; hence it is worthwhile to explore the efficacy of the proposed RP-WNO in learning the mapping between the two. 

In this case study, the mapping is carried out between the initial atomic structure of a two-dimensional porous graphene sheet of size $127.9 \text{\AA}\times 127.8 \text{\AA}$ and the atomic stresses corresponding to a tensile strain of 5\%. Different samples are generated by varying the location of vacancies in the porous graphene sheet randomly. The dataset for this case study is taken from \cite{yang2022linking}, and the readers are advised to follow the same for more details regarding the dataset generation.

\begin{figure}[ht!]
    \centering
    \includegraphics[width = 0.65\textwidth]{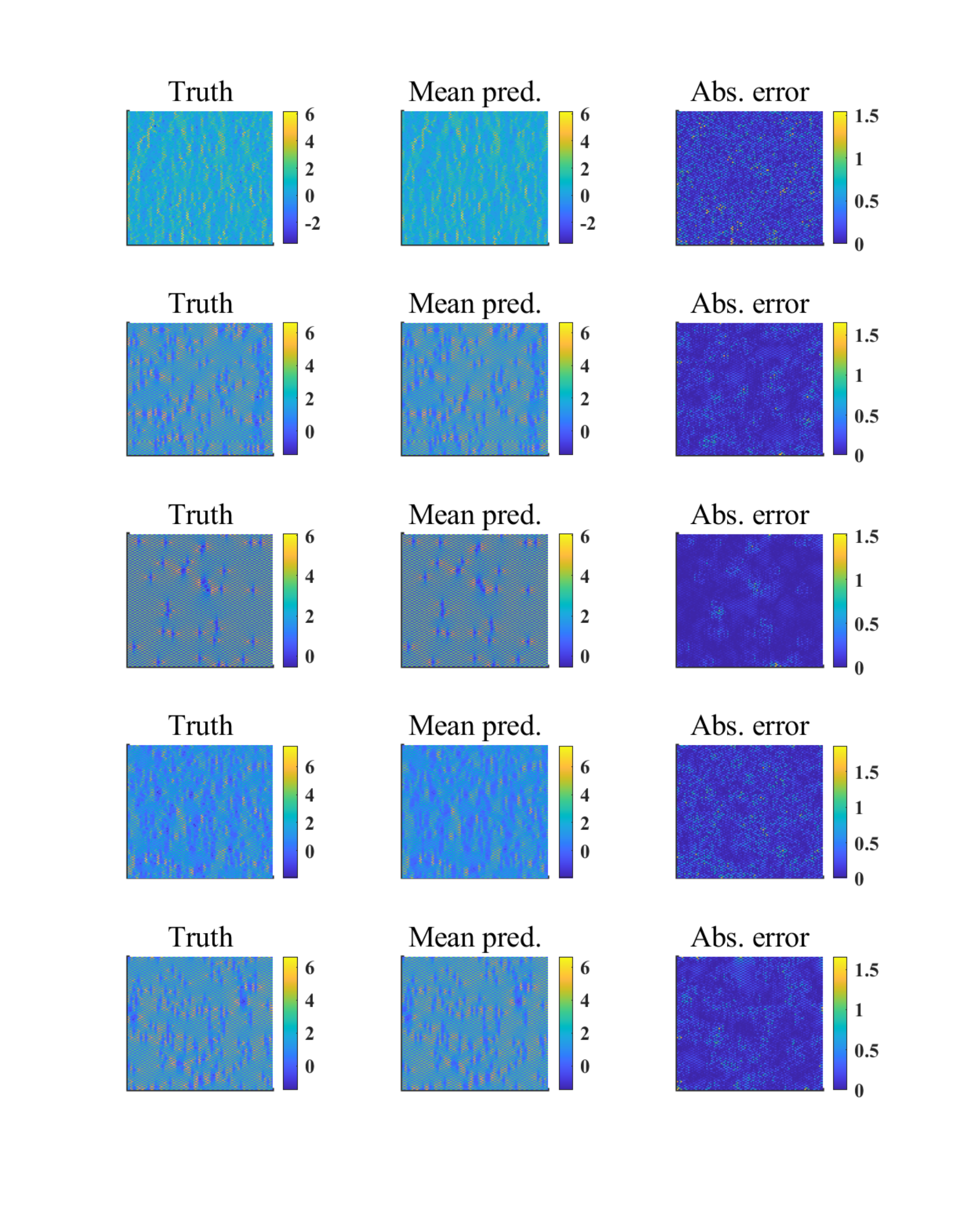}
    \caption{RP-WNO predictions compared against the ground truth for cases study IV.}
    \label{fig: atomic predictions}
\end{figure}
Fig. \ref{fig: atomic predictions} shows the RP-WNO predictions compared against the ground truth. The RP-WNO ensemble was trained using 1500 training samples, and the mean predictions closely follow the ground truth.
\begin{figure}[ht!]
\centering
\begin{subfigure}{0.85\textwidth}
    \centering
    \includegraphics[width = \textwidth]{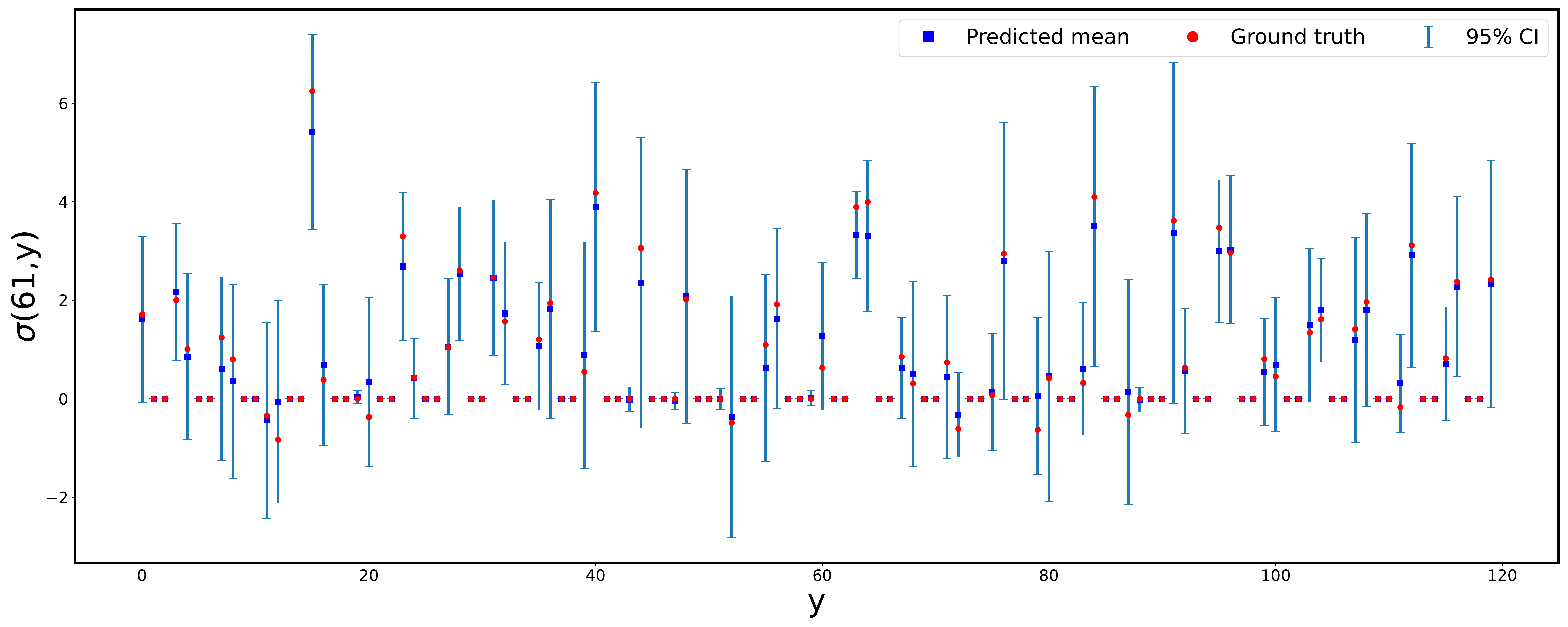}
    \caption{}
\end{subfigure}
\begin{subfigure}{0.85\textwidth}
    \centering
    \includegraphics[width = \textwidth]{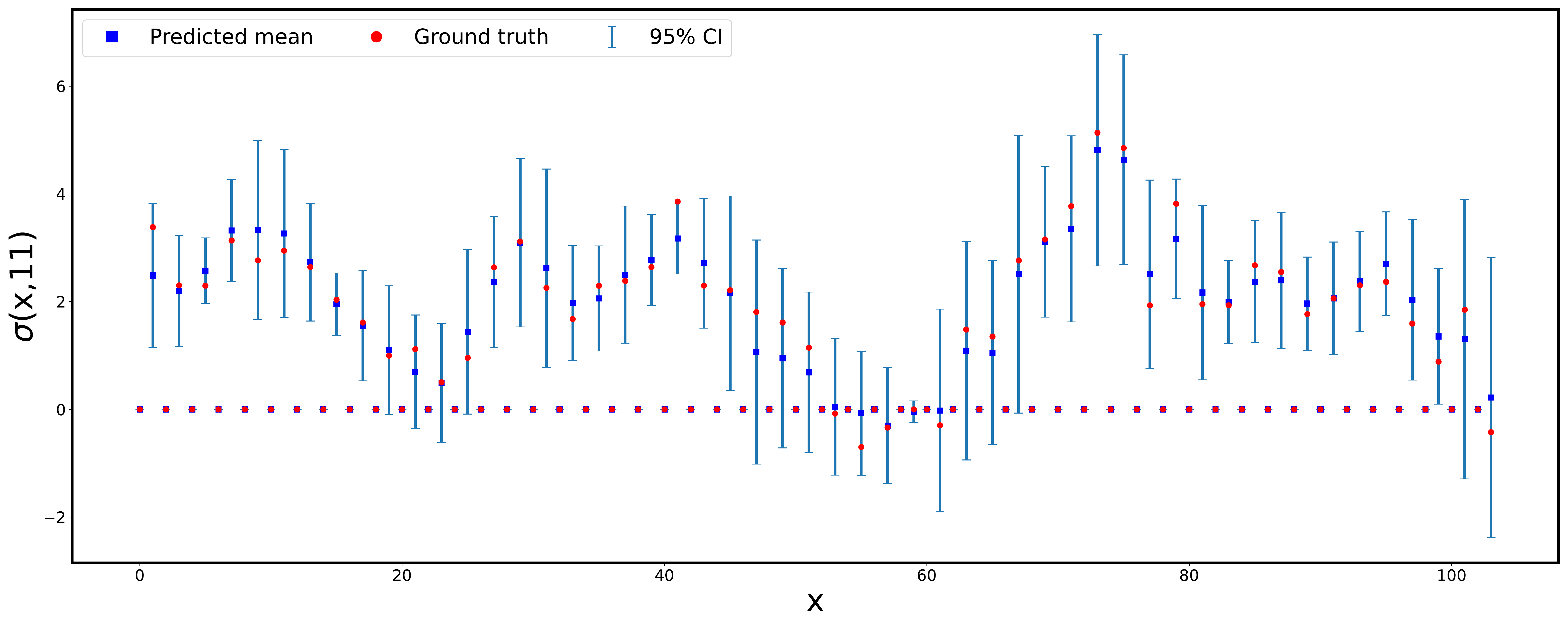}
    \caption{}
\end{subfigure}
    \caption{Plots showcasing the mean prediction and the confidence intervals for one specific initial atomic structure.}
    \label{fig: atomic error bar}
\end{figure}
Fig. \ref{fig: atomic error bar} shows the confidence intervals and mean predictions corresponding to one sample of the prediction ensemble. As can be seen that the mean predictions largely give a good approximation for the ground truth, and for cases where some deviation is observed, the ground truth is almost always encapsulated by the confidence intervals.

\begin{table}[ht!]
\centering
\begin{tabular}{c|c|cccccc|c}\hline
CS-I & CS-II & \multicolumn{6}{c|}{CS-III} & CS-IV \\ \hline
\multirow{3}{*}{0.080} & \multirow{3}{*}{0.095} & \multicolumn{1}{c|}{frequency} & 1Hz & 3Hz & 5Hz & 7Hz & 9Hz & \multirow{3}{*}{2.527} \\\cline{3-8}
 & & \multicolumn{1}{c|}{real part} & 0.082 & 0.100 & 0.250 & 0.674 & 1.918 & \\
 & & \multicolumn{1}{c|}{imaginary part} & 0.002 & 0.003 & 0.005 & 0.008 & 0.013 & \\\hline                    
\end{tabular}
\caption{Percentage NMSE error values for various case studies.}
\label{table: nmse}
\end{table}
\noindent Percentage normalized mean square error values for various case studies are given in Table \ref{table: nmse}.

\section{Conclusion}\label{section: conclusion}
In this paper, we introduce a novel randomized prior wavelet neural operator, which is an operator learning framework and is capable of estimating the uncertainty associated with its predictions. Although Bayesian inference based techniques are often cited when it comes to uncertainty quantification, their complexity and heavy resource requirement can cause hindrances in their augmentation with complex deep learning architectures. The  proposed algorithm gains an advantage here as it is set in the deterministic framework and hence can be trivially implemented for complex deep neural network architectures. The proposed RP-WNO is based on WNO, which utilizes wavelet transform and convolutional layers in order to learn the most relevant information and keep the parameter requirement relatively low for the overall network.

The proposed RP-WNO is trained in an ensemble training setup, which enables uncertainty quantification in the prediction stage. Its advantage over vanilla ensemble training is that by virtue of randomized prior networks, it can encode prior information while training. Uncertainty quantification in the prediction stage can be instrumental for applications where decision-making is involved, and an efficient operator learning framework can by itself be used as a surrogate model in a vast array of applications. To test the proposed framework, four different case studies have been carried out, and the predicted results are compared against the ground truth. The following observations have been made based on the results produced,
\begin{itemize}
    \item The mean predictions in all case studies closely follow the ground truth and give a good approximation for the same.
    \item The training data case studies carried out in CS-I and CS-II show that the width of the confidence intervals obtained using the proposed RP-WNO decrease as the training data is increased and as the mean prediction converges to the ground truth. This is an expected behavior and showcases the ability of the proposed RP-WNO to estimate the true uncertainty associated with its predictions.
    \item The comparison between prediction results obtained from the RP-WNO ensemble utilizing discrete WTs and continuous WTs, in CS-III, shows that even though continuous WT is more resource extensive, it converges better to the ground truth.
    \item Furthermore, while training, it was observed that when continuous WT is used within the RP-WNO architecture, the requirement for the number of nodes in the hidden layer reduces hence compensating slightly for the increased cost of performing continuous WT and learning its various wavelet coefficients.
\end{itemize}
To conclude, based on the results, it can be said that the proposed RP-WNO is able to perform its stipulated tasks efficiently and in an effective manner. Having said all this, the authors note that, like any other research, there is scope for improvement in the proposed RP-WNO framework. For example, research can be carried out to bring down the computational cost associated with training the RP-WNO ensemble. Also, currently, the nature of prior being introduced is not very modular as the architecture for the fixed weights network is kept the same as the trainable network. Hence, the effects of modifying the fixed weight network can be explored in future research. 
\section*{Acknowledgment}
SG acknowledges the financial support received from the Ministry of Education, India, in the form of the Prime Minister's Research Fellows (PMRF) scholarship.
SC acknowledges the financial support received from Science and Engineering Research Board (SERB) via grant no. SRG/2021/000467.
The authors would also like to acknowledge Tapas Tripura, Ph.D. scholar at Applied Mechanics Department, IIT Delhi, for his inputs regarding the continuous wavelet transform in wavelet neural operator.

\appendix
\section{$\beta$ studies}\label{appendix beta studies}
The randomized prior networks require the summation of outputs from the trainable and the non-trainable networks to obtain the final output, using which backpropagation will be carried out for the trainable network. However, before adding the output from the non-trainable network, it is scaled by a factor $\beta$, which is user-defined and can affect the final results produced. In the available literature, \cite{osband2018randomized}, a $\beta$ value of unity is suggested as optimum. To gauge the effect of $\beta$, when randomized prior networks are used in conjunction with WNO, $\beta$ studies are carried out.

The RP-WNO ensemble is trained for the two-dimensional darcy flow equation from CS-II, and 100 samples were taken for training. Multiple RP-WNO ensembles are trained corresponding to different values of $\beta$ parameter, and testing loss corresponding to each ensemble is computed. 

\begin{figure}[ht!]
    \centering
    \includegraphics[width = 0.75\textwidth]{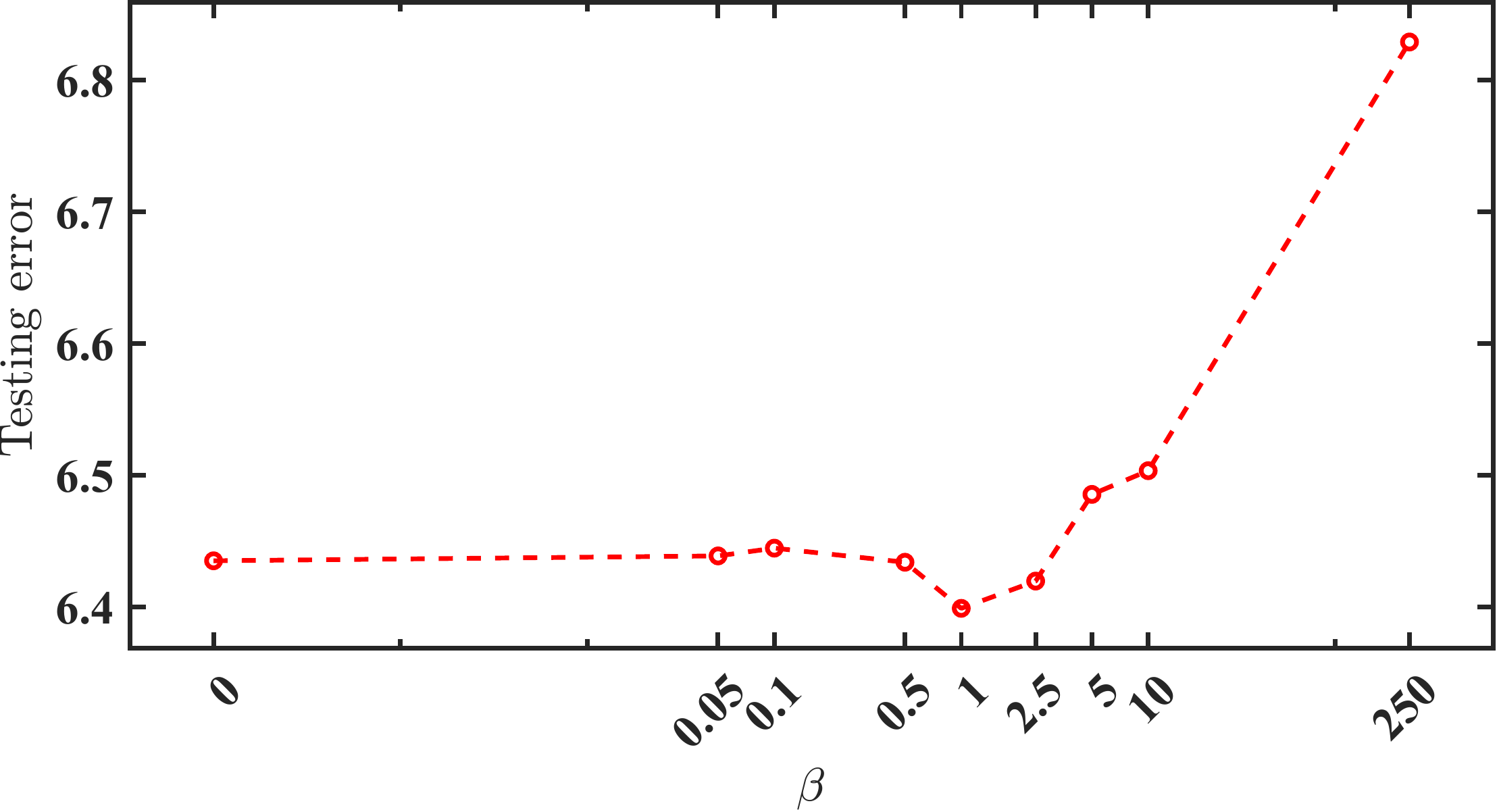}
    \caption{Effect of $\beta$ in the RP-WNO ensemble on the testing loss.}
    \label{fig: beta studies}
\end{figure}
Results shown in Fig. \ref{fig: beta studies} show that a very high value of $\beta$ adversely affects the performance of the RP-WNO ensemble. Furthermore, a very small value of $\beta$ will reduce the effect of randomized prior, thus defeating the purpose of using a randomized prior network. It can be observed from Fig. \ref{fig: beta studies} that the loss stays within a reasonable range for $\beta = 1$, thus reinforcing the findings from the available literature. Authors note that the additional dip in testing error observed near $\beta = 1$ can be chalked up to a statistical anomaly, which may reduce slightly, by changing the random seed used while programming.  
\section{RP-WNO architecture details}\label{appendix RP-WNO}
Architecture details for the RP-WNO ensemble and RP-WNO models will be discussed in this section. 10 copies of RP-WNO models were used for training the RP-WNO ensembles in various case studies. Gaussian error linear unit \cite{hendrycks2016gaussian} was used as activation function, and relative $\mathcal L_2$ error as used in \cite{tripura2022wavelet} is used for training the RP-WNO networks. Details for RP-WNO architectures used in various case studies are as follows.
\subsection{Architecture details CS-I}
The architecture discussed below is for the trainable net, and the same is implemented for the non-trainable net. The WMC of the WNO blocks used in the trainable RP-WNO model of CS-I utilizes discrete WT (Daubechies wavelet \cite{daubechies1992ten}) and performs eight wavelet transforms to distill the relevant information. The wavelet coefficients obtained after the final WT are learned using the linear dense layers. Coefficients obtained in the previous WTs are used as is while inverse WT. Furthermore, inside the CMC of WNO blocks, a single convolution layer is used with a kernel size of one, to learn the inputs of CMC. Four such WNO blocks are used in the trainable RP-WNO model.

Now, the first dense layer used to uplift the trainable RP-WNO input has 64 nodes. The final forward neural net after the fourth WNO block has two dense layers with 128 nodes in the first layer and a single node in the last output layer. GeLU activation is used after the first, second, and third wavelet block and between the two dense layers of the forward neural net at the end.

\subsection{Architecture details CS-II}
Similar to CS-I, WMC in CS-II utilizes discrete WT (Daubechies wavelet) and performs four WTs to distill the relevant information. The approximate and horizontal wavelet coefficients obtained after the final WT are learned using the linear dense layers. Coefficients obtained in the previous WTs and the ones which are not learned are used as is while inverse WT. Similar to architecture in CS-I, CMC of WNO blocks use a single convolution layer with a kernel size of one, to learn its inputs. Four such WNO blocks are used in the trainable RP-WNO model.

The first dense layer used to uplift the trainable RP-WNO input has 64 nodes, and the final forward neural net has two dense layers with 192 nodes in the first layer and a single node in the last output layer. GeLU activation is used after the first, second, and third wavelet block and between the two dense layers of the forward neural net at the end.

\subsection{Architecture details CS-III}
WMC in CS-III utilizes continuous WT and performs four WTs to distill the relevant information. All wavelet coefficients obtained after the final WT are learned using the linear dense layers. Coefficients obtained in the previous WTs are used as is while inverse WT. Similar to previous architectures, the CMC of WNO blocks uses a single convolution layer with a kernel size of one, to learn its inputs. Four such WNO blocks are used in the trainable RP-WNO model.

The first dense layer used to uplift the trainable RP-WNO input in CS-III has 64 nodes, and the final forward neural net has two dense layers with 64 nodes in the first layer and a single node in the last output layer. GeLU activation is used after the first, second, and third wavelet block and between the two dense layers of the forward neural net at the end. The architecture used to train the RP-WNO ensemble learning the real part of the pressure wavefield is kept the same as the one used for learning the imaginary part of the pressure wavefield.

\subsection{Architecture details CS-IV}
WMC in CS-IV utilizes continuous WT as in CS-III and performs three WTs to distill the relevant information. All wavelet coefficients obtained after the final WT are learned using the linear dense layers. Coefficients obtained in the previous WTs are used as is while inverse WT. Similar to previous architectures, the CMC of WNO blocks uses a single convolution layer with a kernel size of one, to learn its inputs. Five such WNO blocks are used in the trainable RP-WNO model.

The first dense layer used to uplift the trainable RP-WNO input has 128 nodes, and the final forward neural net has two dense layers with 192 nodes in the first layer and a single node in the last output layer. GeLU activation is used after the first, second, third, and fourth wavelet block and between the two dense layers of the forward neural net at the end.

\end{document}